\documentclass[10pt,twocolumn,letterpaper]{article}

\usepackage{cvpr}              

\usepackage{pifont}
\usepackage{multirow}
\usepackage{booktabs}
\usepackage{arydshln} 
\usepackage{colortbl}  
\PassOptionsToPackage{table,xcdraw}{xcolor}
\usepackage{xcolor}
\usepackage{amsmath}
\usepackage{float}
\newcommand{\cmark}{\ding{51}} 
\newcommand{\xmark}{\ding{55}} 

\definecolor{cvprblue}{rgb}{0.21,0.49,0.74}
\usepackage[pagebackref,breaklinks,colorlinks,allcolors=cvprblue]{hyperref}

\def\paperID{5782} 
\def\confName{CVPR}
\def\confYear{2026}

\title{Continual Alignment for SAM:  Rethinking Foundation Models for Medical Image Segmentation in Continual Learning}

\author{
    Jiayi Wang\textsuperscript{1}\thanks{Equal contribution.},\hspace{0.2cm}
    Wei Dai\textsuperscript{1}\footnotemark[1],\hspace{0.2cm}
    Haoyu Wang\textsuperscript{1}\footnotemark[1],\hspace{0.2cm}
    Sihan Yang\textsuperscript{1}\thanks{Project leader.},\hspace{0.2cm}
    Haixia Bi\textsuperscript{2}\thanks{Corresponding author. \ Email: {\ haixia.bi@xjtu.edu.cn}}, \hspace{0.2cm}
    Jian Sun\textsuperscript{3},     \vspace{3pt}\\
    \fontsize{11}{13}\selectfont \textsuperscript{1}Xi'an Jiaotong University \hspace{0.2cm}\\
    \fontsize{11}{13}\selectfont \textsuperscript{2}School of Information and Communications Engineering, Xi'an Jiaotong University \hspace{0.2cm}\\
    \fontsize{11}{13}\selectfont \textsuperscript{3}School of Mathematics and Statistics, Xi'an Jiaotong University
}

\begin{document}
\maketitle
\begin{abstract}
In medical image segmentation, heterogeneous privacy policies across institutions often make joint training on pooled datasets infeasible, motivating continual image segmentation—learning from data streams without catastrophic forgetting. While the Segment Anything Model (SAM) offers strong zero-shot priors and has been widely fine-tuned across downstream tasks, its large parameter count and computational overhead challenge practical deployment. 
This paper demonstrates that the SAM paradigm is highly promising once its  computational efficiency and performance can be balanced.
To this end, we introduce the \textbf{Alignment Layer}, a lightweight, plug-and-play module which aligns encoder–decoder feature distributions to efficiently adapt SAM to specific medical images, improving accuracy while reducing computation. 
Building on SAM and the Alignment Layer, we then propose \textbf{C}ontinual \textbf{A}lignment for \textbf{SAM} (\textbf{CA-SAM}), a continual learning strategy that automatically adapts the appropriate Alignment Layer to mitigate catastrophic forgetting, while leveraging SAM’s zero-shot priors to preserve strong performance on unseen medical datasets.
Experimented across nine medical segmentation datasets under continual-learning scenario, CA-SAM achieves state-of-the-art performance. 
Our code, models and datasets will be released on \mbox{https://github.com/azzzzyo/Continual-Alignment-for-SAM.}
\end{abstract}

\section{Introduction}
\label{sec:intro}

\begin{figure}[t]
  \centering
  \includegraphics[width=\linewidth]{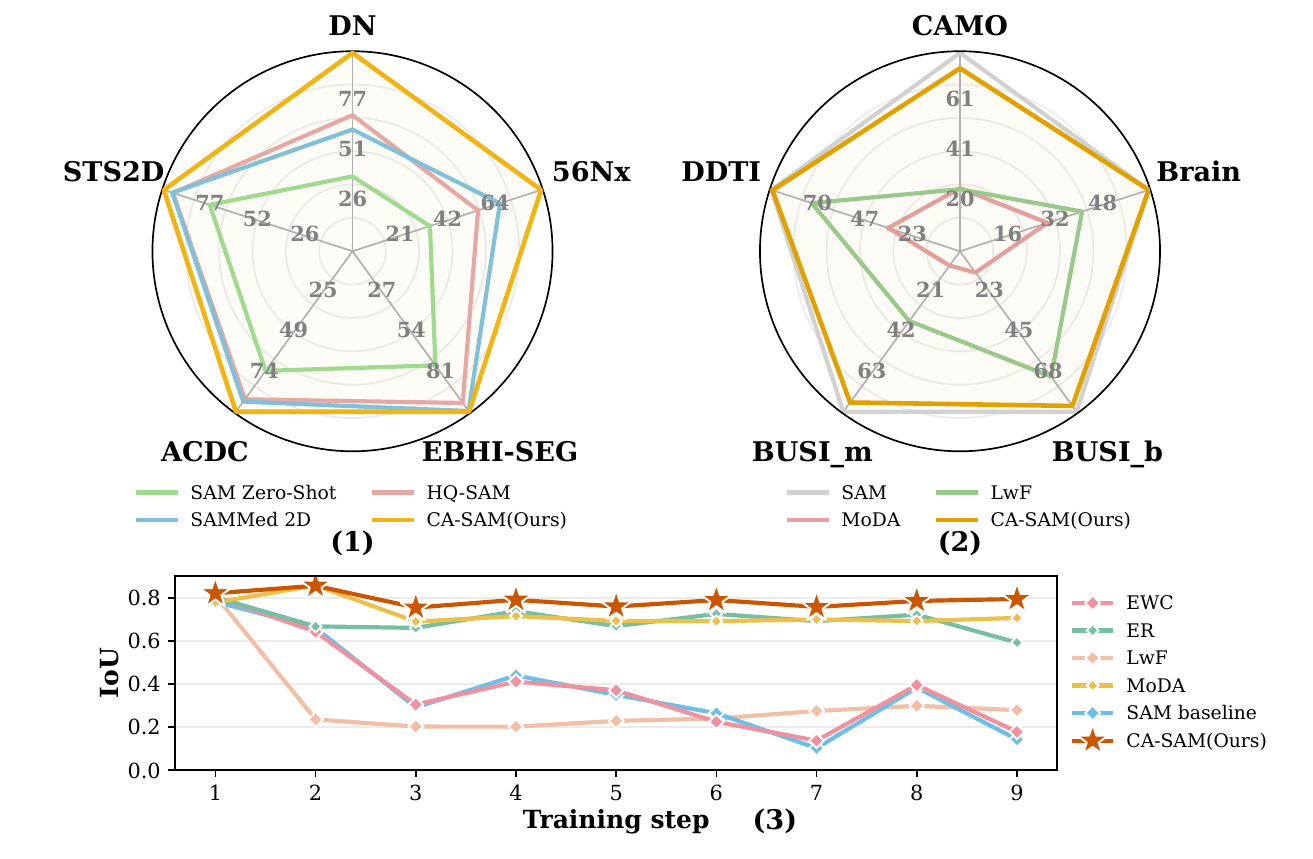}
  \caption{Comparison with prior methods across three aspects. We compare \textbf{CA-SAM (our method)} against baselines in: (1) Single-dataset segmentation performance (Radar Plot), (2) Zero-shot performance on unseen datasets (Radar Plot), and (3) Cross-dataset continual segmentation performance (Line Chart).}
  \label{fig:intro_fig}
\end{figure}

In medical image segmentation, privacy and governance constraints across institutions often preclude joint training on pooled datasets, making continual segmentation—learning from sequential data streams without revisiting prior samples—practically essential yet technically challenging.
Meanwhile, the Segment Anything Model (SAM~\cite{kirillov2023segment}) has demonstrated strong segmentation performance and \textit{zero-shot} capability. Its potential in continual learning settings warrants further exploration~\cite{yang2024continual,cheng2023sammed2d,zhu2023continual}.

However, the parameter scale and computation footprint of SAM make efficient adaptation non-trivial. This bottleneck enlightens our first exploratory question: 
Under realistic annotation budget constraints and computation limits, do SAM-based adaptation methods offer a better cost–benefit ratio than robust CNN pipelines in practice?

Our quantitative study shows that full-parameter fine-tuning on SAM significantly outperforms CNNs but incurs a substantial increase in computational overhead.
Parameter-efficient fine-tuning markedly reduces trainable parameters and computation. 
Nevertheless, it performs only on par with CNN pipelines~\cite{ronneberger2015U-Net,zhou2018unet++}, failing to fully leverage the inherent advantages of SAM.
Furthermore, in the context of Continual Learning, SAM-based adaptation methods designed for downstream tasks often erode SAM’s broad generalization capabilities, particularly for unseen domains.

Collectively, the evidence motivates us to rethink how SAM should be adapted for continual learning in medical image segmentation: rather than maximizing in-domain accuracy at all costs, we emphasize the importance of computational efficiency and generalization.

Concretely, we identify three limitations in current SAM-based approaches:(1) They struggle to balance computational efficiency and performance~\cite{cheng2023sammed2d,chen2023sam}, as they require substantial parameter updates to the backbone networks; (2) they typically depend on stored exemplars or rehearsal~\cite{zhu2023continual,wang2023rethinking}, violating exemplar-free constraints common in clinical settings; (3) after fine-tuning $\text{\cite{wu2025medical,shi2025segment}}$, they tend to erode SAM’s generalization, yielding poor out-of-distribution (OOD) performance.

In response, we introduce a lightweight, plug-and-play \textbf{Alignment Layer} inserted between the frozen SAM encoder and decoder to align latent feature distributions within the target medical domain. By confining backpropagation to this module, we adapt SAM to dataset-specific statistics while preserving the capacity of the original backbone and substantially reducing computation.
Building on this~module, we propose \textbf{C}ontinual \textbf{A}lignment for \textbf{SAM} (\textbf{CA-SAM}), a novel continual learning framework implemented via two core components: task-specific Alignment Layers and an exemplar-free task routing mechanism. The latter, which is realized via task-calibrated Variational Autoencoders (VAEs), serves a dual purpose: it automatically adapts  the Alignment Layer per task to mitigate catastrophic forgetting and falls back to the frozen SAM for zero-shot inference on unseen domains. 
Furthermore, CA-SAM exhibits strong robustness to task variations, a desirable virtue for real-world medical application scenarios where data patterns are inherently unpredictable.
This design ensures both task-specific plasticity and SAM's broad generalization, significantly enhancing continual learning performance 
in medical image segmentation.

We evaluate our CA-SAM across nine medical segmentation datasets spanning multiple organs and modalities. 
Among SAM-based partial-fine-tuning methods~\cite{cheng2023sammed2d,sam_hq}, our CA-SAM achieves state-of-the-art efficiency–performance trade-offs in computation, trainable parameters and accuracy. 
Under realistic streaming protocols, CA-SAM delivers strong performance with substantially reduced forgetting. 
Moreover, after medical-task training, the VAE Router preserves near-upper-bound zero-shot performance on previously unseen domains by correctly routing OOD inputs back to frozen SAM. Together, these results establish CA-SAM as a practical and effective recipe for continual learning in medical image segmentation with foundation models. \Cref{fig:intro_fig} presents the evaluation results of CA-SAM against other methods across three aspects: single-dataset segmentation performance, zero-shot capability and \mbox{cross-dataset stability.}

In summary, our contributions are as follows:

\begin{itemize}
    \item We propose Alignment Layer: a lightweight adapter between the frozen encoder and decoder, which aligns latent representations to the medical target domain, delivering favorable computation–performance trade-offs.
    \item We propose CA-SAM: an exemplar-free continual learning framework with VAE-based task routing and OOD fallback to frozen SAM, mitigating forgetting while preserving zero-shot generalization.
    \item  We conduct a comprehensive evaluation: evidence across nine heterogeneous datasets, realistic streaming settings and explicit OOD tests demonstrating state-of-the-art efficiency and accuracy.
\end{itemize}

\section{Related Work}
\label{sec:Related Work}
\subsection{SAM for Medical Image Segmentation}


The domain gap between natural and medical images is the most critical factor which limits SAM's performance in medical scenarios. Previous researchers have proposed several improvement strategies for pretrained SAM~\cite{cheng2023sam,shaharabany2023autosam,cheng2024unleashing}. For example, SAM-Med2D~\cite{cheng2023sammed2d}, SAMed~\cite{yan2025samed} and Medical SAM Adapter~\cite{wu2025medical} tried to fine-tune SAM using multiple methods including LoRA and domain specific Adapter; SyncSAM ~\cite{yang2025improved} introduced a convolutional branch in parallel and used multi-scale feature fusion in the mask decoder; Self-Prompt SAM~\cite{xie2025self} and Semi-Supervised SAM-2~\cite{zhu2025sss}, respectively, employ a multi-scale prompt generator and a discriminative enhancement mechanism to enhance SAM’s generalization in the medical domain.
Our method, while retaining the strong generalization of pretrained SAM, efficiently performs transfer learning at the feature level using Alignment Layer with significantly fewer parameters, directly finding the best distribution for each dataset.

\subsection{Segmentation Tasks in Continual Learning}
Before the emergence of SAM, several studies had explored the application of continual learning in medical image segmentation. These methods can be broadly categorized into three paradigms: replay-based~\cite{bera2023memory,rolnick2019experience,isele2018selective,lopez2017gradient}, regularization-based~\cite{kirkpatrick2017overcoming,li2017learning,aljundi2018memory}, and dynamic-expansion~\cite{aljundi2017expert,yoon2017lifelong,yan2021dynamically,douillard2022dytox,hu2023dense} approaches. With the emergence of SAM, continual learning has entered a new phase. Using SAM strong segmentation ability and parameter-efficient fine-tuning (PEFT) techniques, recent studies pursue efficient task adaptation. Existing methods mainly include dynamic structural extensions~\cite{yang2024continual,chen2023sam,wu2025medical}, feature or parameter adaptation~\cite{shu2025regcl,shi2025segment,Fukuda_2025_CVPR}, and memory-based strategies~\cite{cheng2023sammed2d,zhu2023continual,wang2023rethinking}. However, these approaches often suffer from inefficiency, loss of task-specific information, and the lack of mechanisms to handle unseen domains, which leads to degraded generalization of SAM. To address these limitations, we propose \textbf{CA-SAM}, a lightweight continual learning framework with learnable alignment layers and a VAE-based task routing mechanism.
This design effectively preserves historical knowledge and mitigates zero-shot degradation, enabling efficient continual learning in medical image segmentation while retaining SAM’s strong generalization.


\section{Method}
\label{sec:Method}

\subsection{Do SAM-Based  Gains Offset Resource Costs?}

\begin{figure}[h]  
  \centering  
  \includegraphics[width=0.9\linewidth]{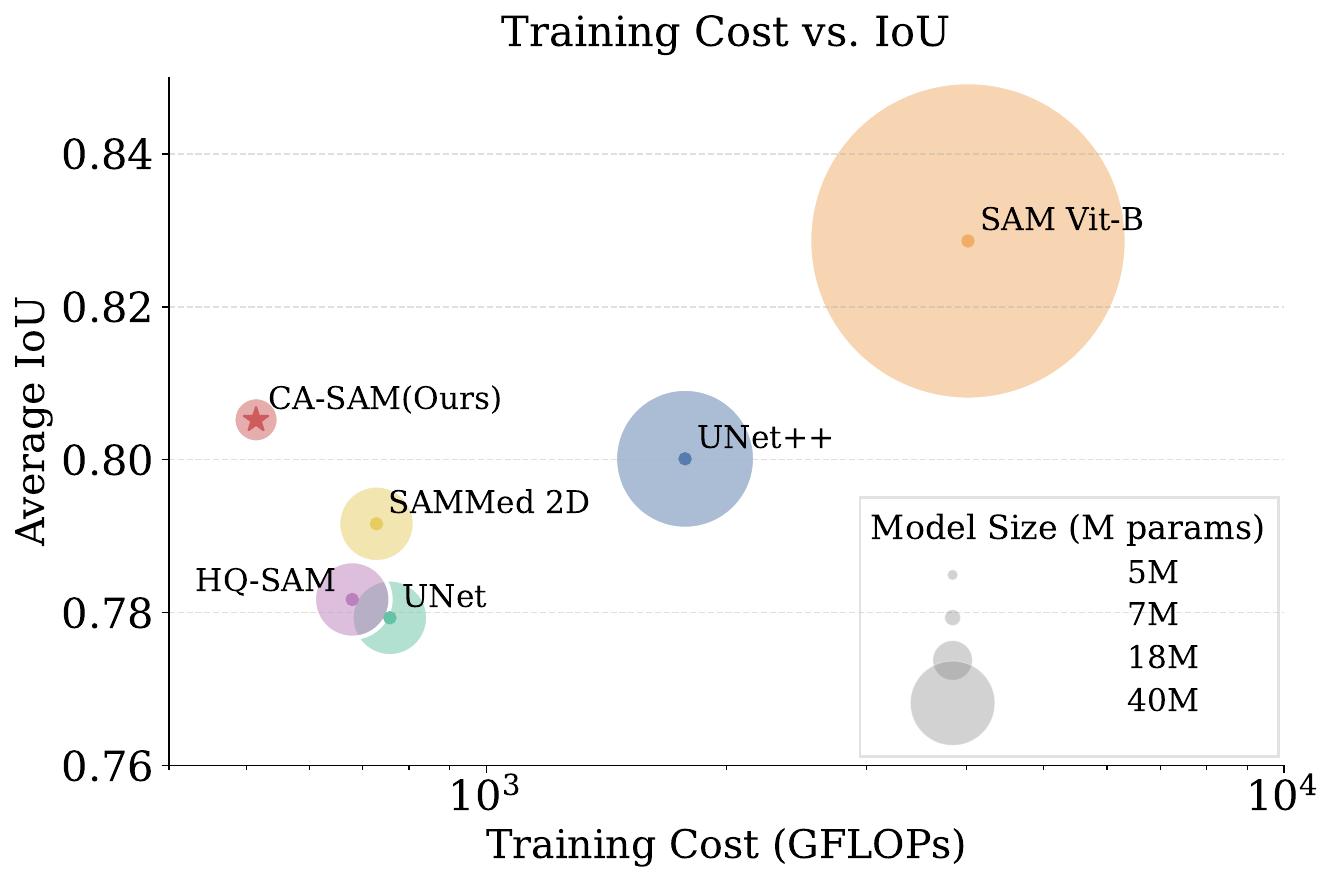}  
  \caption{A comprehensive comparison of IoU scores, training cost, and trainable parameters (model size).
The training computation cost is measured as the 
FLOPs incurred by one training pass on a standard-size image (3×1024×1024).}
  \label{fig:IoU-cost-trainable_parameters}
\end{figure}

In recent years, SAM has been increasingly applied to medical image segmentation owing to its strong zero-shot segmentation capability. However, fine-tuning SAM, particularly full parameter fine-tuning, often entails large parameter counts and sharply increased computational costs. This raises a natural question: \textbf{are the performance gains of SAM-based methods sufficient to offset these resource costs?} To investigate this, we conduct a quantitative comparison on three medical datasets~\cite{cheng2025interactive}, contrasting CNN-baseds~\cite{ronneberger2015U-Net,zhou2018unet++} and SAM-based methods in terms of IoU/BIoU, computational cost, and the number of trainable parameters. The results are presented in \Cref{fig:IoU-cost-trainable_parameters}.

Our quantitative analysis shows that parameter-efficient SAM-based fine-tuning substantially reduces computational costs and the number of trainable parameters, yields clear gains in Boundary IoU(BIoU), and maintains IoU that is broadly comparable to CNN-based baselines. Meanwhile, the pronounced improvements observed with full parameter fine-tuning indicate that SAM-based approaches retain strong performance potential for medical image segmentation. Nevertheless, the parameter and computational demands of full parameter fine-tuning remain substantial. Therefore, improving fine-tuning efficiency and reducing training resource consumption are essential for the real-world deployment of SAM in this domain, which motivates our exploration of lightweight, parameter-efficient designs.


\subsection{Alignment Layer}

Model parameter fine-tuning methods aim to adapt feature representations across domains through distribution alignment.
Since each medical dataset exhibits a distinct feature distribution, we assign a dedicated Alignment Layer to each medical dataset to align its features to an optimal distribution subspace.

Given an input image $I$, the structure of the pre-trained SAM can be formulated as:
\begin{equation}
    Y = D(E(I)), \quad I \in \mathbb{R}^{B \times 3 \times H \times W},
    \label{the_structure_of_the_pre-trained_SAM}
\end{equation}
where \(E(\cdot)\)  and \(D(\cdot)\)  denote the frozen Encoder and Decoder of SAM, respectively.

Considering the large number of parameters in the pre-trained SAM, we freeze both \(E(\cdot)\) and \(D(\cdot)\), and introduce a lightweight trainable module $\mathcal{A}(\cdot)$ between them to directly adapt the latent feature representation $Z$:
\begin{equation}
    \qquad Z = E(I),\quad \mathbf{\tilde{Z}} = \mathcal{A}(Z).
    \label{adapt_the_intermediate_feature}
\end{equation}

This module $\mathcal{A}(\cdot)$ progressively approximates the target feature distribution by stacking multiple alignment layers, 
which are defined as several convolutional layers. The model architecture is illustrated in \Cref{sec:alignment_layer}.
\begin{equation}
    Y = D(\mathbf{\tilde{Z}}) = D\!\left(\mathcal{A}(E(I))\right), \quad Y \in \mathbb{R}^{B \times 1 \times H \times W}.
    \label{The_model_architecture}
\end{equation}

This design introduces a minimal number of trainable parameters, with backpropagation restricted to the Alignment Layer, leaving the frozen Encoder and Decoder unaffected.
Our method significantly reduces computational cost while preserving the architecture and capabilities of the pre-trained SAM, enabling efficient and stable domain adaptation compared to other fine-tuning approaches.

\begin{figure*}[t]  
  \centering  
  \includegraphics[width=1.0\linewidth]{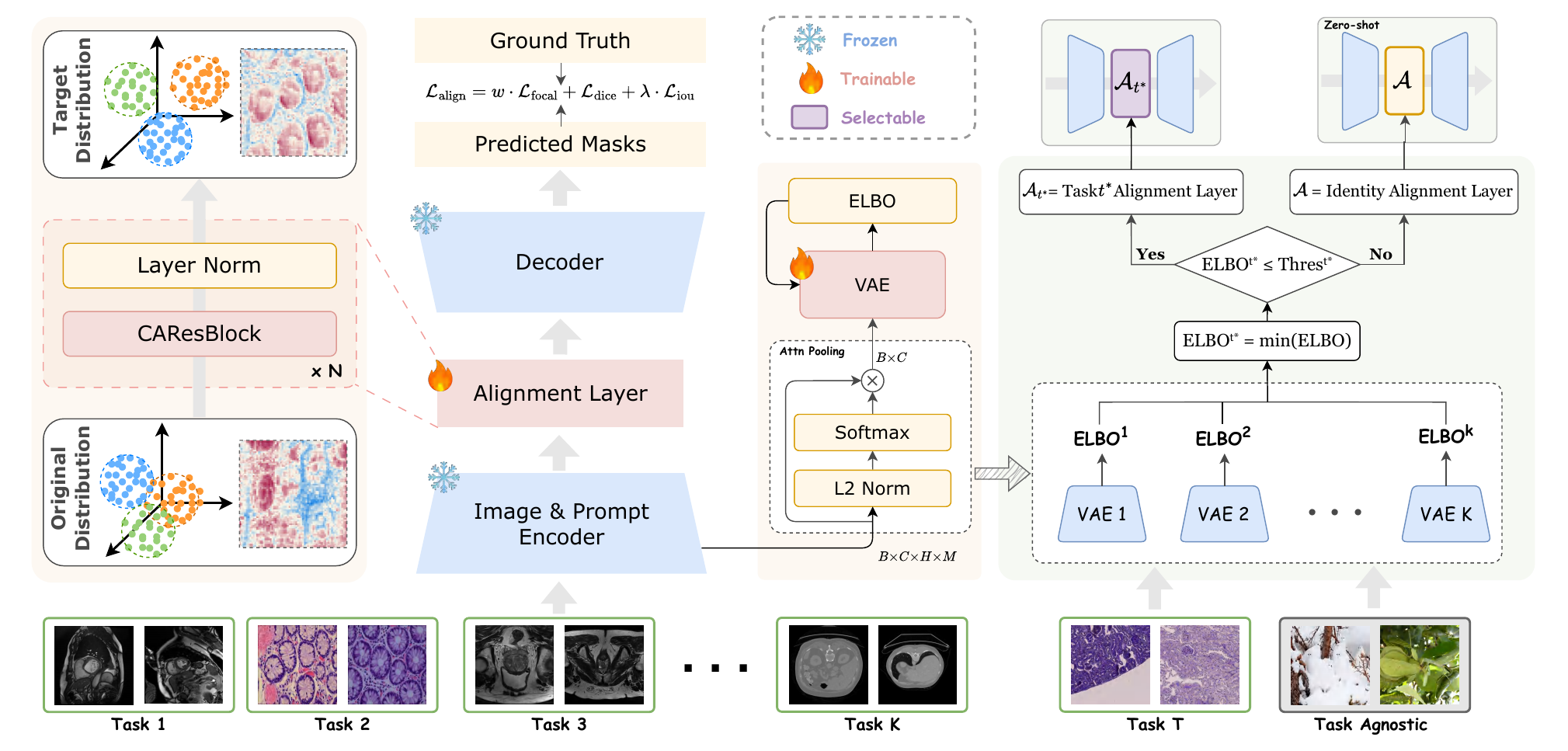}  
  \caption{\textbf{Framework of Continual Alignment for SAM.} The figure shows, from left to right, the structure of the alignment layer, backbone architecture, the training procedure of the VAE, and the CA-SAM routing mechanism along with its OOD fallback mechanism.}
  \label{fig:Framework_of_CA-SAM}
\end{figure*}

\subsection{Continual Alignment for SAM}

Due to privacy constraints in medical data, joint training across sources is often infeasible.  
To enable low-forgetting medical image segmentation under sequential training, we propose CA-SAM.  
Built upon a lightweight alignment layer, CA-SAM introduces an exemplar-free task routing mechanism that enables stable task identification and OOD fallback while preserving SAM’s generalization ability, as illustrated in \Cref{fig:Framework_of_CA-SAM}.

\noindent\textbf{Continual Learning Setting.} In continual learning, training data arrive sequentially as:
\begin{equation}
    D^{tr}=\{D^{tr}_1,\ldots,D^{tr}_N\},\quad D^{te}=\{D^{te}_1,\ldots,D^{te}_N\},
    \label{training_data_arrive_sequential}
\end{equation}
where \(D^{tr}_t\) and \(D^{te}_t\) denote the training and testing sets of the \(t\)-th task, and \(N\) is the total number of tasks.  
At each stage \(t\), the model can only access the current dataset \(D^{tr}_t\) without revisiting previous samples.  
After training, evaluation is performed on the accumulated test sets \(\{D^{te}_1,\ldots,D^{te}_t\}\) to measure both current-task performance and the degree of forgetting using IoU and BIoU metrics.

\noindent\textbf{Task Routing Mechanism.}
For medical image segmentation tasks exhibit significant differences in feature distribution and semantic structure,  
we assign each task \(t\) a dedicated Alignment Layer \(\mathcal{A}_t(\cdot)\) and train an independent VAE \(\mathcal{V}_t = (\text{Enc}_t, \text{Dec}_t)\)  
to model the probabilistic feature distribution \(p_t(f)\) of that task.  
Given the frozen encoder feature map \(Z\), a global feature vector \(f \in \mathbb{R}^D\) is obtained through a parameter-free attention pooling mechanism:
\begin{equation}
    \alpha_{h,w} = \mathrm{softmax}\!\left(\frac{\|Z_{:,h,w}\|_2}{C \cdot T}\right),  
    f = \sum_{h,w} \alpha_{h,w} Z_{:,h,w},
    \label{parameter-free_attention_pooling_mechanism}
\end{equation}
where \(T\) is a temperature coefficient, and \(C\) denotes the number of feature channels of \(Z\).
This process measures spatial saliency via the L2 norm and aggregates local features according to their importance.  

Each task-specific VAE \(\mathcal{V}_t\) is optimized by minimizing the evidence lower bound (ELBO) loss:
\begin{equation}
    \mathcal{L}_{VAE}(f) =
    \tfrac{1}{D}\|f - \hat{f}\|_2^2 +
     \tfrac{\beta}{2} \sum_{i=1}^{D} (\mu_i^2 + \sigma_i^2 - 1 - \log\sigma_i^2),
     \label{ELBO_loss}
\end{equation}
where the first term is the reconstruction loss preserving feature consistency,  
and the second term is the Kullback–Leibler (KL) divergence regularization between the latent distribution
\(q_t(z|f)=\mathcal{N}(\mu,\sigma^2)\) and the standard normal prior \(p(z)=\mathcal{N}(0,I)\).  
The coefficient \(\beta\) balances these two terms, where a larger \(\beta\) enforces stronger latent regularization,  
guiding the model to focus on high-level semantic distributions that better capture the inter-task differences in medical images.  
After optimization, each \(\mathcal{V}_t\) outputs a task similarity score \(s_t(f)\), which is used for task discrimination during inference.

\noindent\textbf{Task Discrimination and Fallback.}
During inference, CA-SAM employs the trained VAEs to automatically determine the most probable task for a given input.  
The feature representation $f$ is passed through all task-specific VAEs to compute ELBO-based scores:
\begin{equation}
    s_t(f)=\mathcal{L}^{(t)}_{\text{VAE}}(f), \quad t^* = \arg\min_t s_t(f).
    \label{compute_ELBO-based_scores}
\end{equation}

The task with the lowest score is regarded as the most likely match. 
To enhance stability, a confidence threshold $\tau_t$ is calibrated for each task via $K$-fold cross-validation on the training set, using the ELBO distribution of the \mbox{held-out folds.}

If $s_{t^*}(f) \le \tau_{t^*}$, the model loads the corresponding alignment layer $\mathcal{A}_{t^*}$ for segmentation.
Otherwise, the input is considered OOD and does not belong to any learned task.  
In this case, CA-SAM employs an identity alignment layer $\mathcal{A}_{id}(Z) = Z$, which reverts to the frozen SAM for zero-shot segmentation inference.
This VAE Router and fallback strategy enables robust task identification and dynamic routing without explicit task labels.
\section{Experiment}
\label{sec:Experiment}
\definecolor{mycrimson}{HTML}{DC143C}   
\definecolor{mylavender}{HTML}{E6E6FA}  
\begin{table*}[!t]
  \centering
  \begin{minipage}{\textwidth}
  \centering
  \setlength{\tabcolsep}{4pt}
  \renewcommand{\arraystretch}{1.15}

  \begin{tabular}{l|c||ccc|ccc}
    \toprule\toprule
    \multirow{2}{*}{\textbf{Method}} & \multirow{2}{*}{\textbf{EF}} &
    \multicolumn{3}{c|}{\textbf{IoU on Med}} & \multicolumn{3}{c}
    {\textbf{BIoU on Med}}\\
           & & Last-IoU \(\uparrow\) & Avg-IoU \(\uparrow\) & FF-IoU \(\downarrow\)
             & Last-BIoU \(\uparrow\) & Avg-BIoU \(\uparrow\) & FF-BIoU \(\downarrow\) \\
    \midrule
    Joint training                                & \cmark & 78.43 & —      & —      & 61.26 & —      & —      \\
    SAM~\cite{kirillov2023segment}                & \cmark & 53.81 & —      & —      & 36.52 & —      & —      \\
    \midrule
    SAM~\cite{kirillov2023segment} + AL(naive)     & \cmark & 14.38 & 37.93 & 65.79\% & 13.50 & 29.91 & 49.69\% \\
    LwF~\cite{li2017learning}                     & \cmark & 27.84 & 30.77 & 3.03\%   & 22.82 & 28.45 & 3.05\%  \\
    EWC~\cite{kirkpatrick2017overcoming}          & \cmark & 17.78 & 38.51 & 57.32\%  & 12.58 & 30.42 & 42.16\% \\
    EMR~\cite{huang2024emr}          & \cmark & 43.09 & 53.96 & 10.95\%  & 26.05 & 33.77 & 8.53\% \\
    ER~\cite{rolnick2019experience}               & \xmark & 59.19 & 69.59 & 21.68\%  & 39.93 & 51.37 & 24.25\% \\
    DER~\cite{buzzega2020darkexperiencegeneralcontinual} & \xmark & 59.99 & 63.14 & 6.55\%   & 40.69 & 42.20 & 6.69\%  \\
    L2P~\cite{wang2022learning}                   & \cmark & 57.34 & 53.84 & 1.55\%   & 37.86 & 35.70 & 1.79\%  \\
    MoDA~\cite{yang2024continual} + HQ-SAM~\cite{sam_hq} & \xmark
      & 69.30 & 70.86 & 2.00\% 
    & \cellcolor{mylavender!50}{55.52}
    & \cellcolor{mylavender!50}{55.88}
    & 1.64\% \\

    MoDA~\cite{yang2024continual}                 & \xmark
      & \cellcolor{mylavender!50}{70.64}
      & \cellcolor{mylavender!50}{72.44}
      & \cellcolor{mycrimson!18}{0.66\% }
      & 52.12 & 54.45 & \cellcolor{mylavender!50}{0.65\%}  \\
    CA-SAM (Ours)                                 & \cmark
      & \cellcolor{mycrimson!18}\(\mathbf{76.12}\)
      & \cellcolor{mycrimson!18}\(\mathbf{76.90}\)
      & \cellcolor{mylavender!50}\(\textbf{1.43\% }\)
      & \cellcolor{mycrimson!18}\(\mathbf{59.95}\)
      & \cellcolor{mycrimson!18}\(\mathbf{59.45}\)
      & \cellcolor{mycrimson!18}\(\mathbf{0.24\%}\) \\
    \bottomrule\bottomrule
  \end{tabular}

  \caption{
  \textbf{Continual Learning results on medical datasets.}  
  ``EF'' indicates Exemplar-free continual learning methods that do not access previous task samples during current task training.
  All compared methods adopt an Alignment Layer for SAM adaptation, except \textit{MoDA + HQ-SAM}, which uses the HQ-SAM decoder. 
  Details of metrics are in Appendix.
  The
    \colorbox{mycrimson!18}{best}
     and 
    \colorbox{mylavender!50}{second best}
    performances are highlighted.
  }
  \label{tab:continual-learning-results-med}
  \end{minipage}
\end{table*}

\subsection{Experimental Settings}

\subsubsection{Datasets}

To evaluate our method across diverse clinical settings, we use nine medical image segmentation datasets spanning organs (heart, spleen, prostate, kidney, and teeth) and modalities—magnetic resonance (MR), computed tomography (CT), and histopathology.
The datasets are \textbf{ACDC}~\cite{cheng2025interactive}, \textbf{EBHI-SEG}~\cite{shi2023ebhi},  \textbf{56Nx}~\cite{tang2024holohisto},  \textbf{DN}~\cite{tang2024holohisto},  \textbf{Polyp}~\cite{jha2019kvasir},  \textbf{MSD\_Prostate}~\cite{cheng2025interactive},  \textbf{MSD\_Spleen}~\cite{cheng2025interactive},  \textbf{Promise12}~\cite{cheng2025interactive}, and  \textbf{STS-2D}~\cite{wang2025sts2d}. For each dataset, we follow the official train and test split. 


\subsubsection{Implementation Details}

\textbf{General Configuration.}  
All experiments are conducted on a single NVIDIA RTX A5000 GPU (24 GB). 
All input images are uniformly resized to $1024\times1024$ to ensure compatibility with the original SAM architecture. 
The Adam optimizer is adopted with an initial learning rate of $1\times10^{-4}$, a batch size of 6 and the number of training epochs 24 for each single-task experiment. 
During training, random points or boxes are selected as prompts, while during testing, only box prompts are employed. 
The segmentation performance is evaluated on the cumulative test set of all seen tasks using two metrics: IoU and BIoU.\\
\textbf{Alignment Layer Configuration.}  
\label{sec:alignment_layer}
Specifically, our Alignment Layer consists of several stacked Channel Attention Residual Blocks (CAResBlock), each designed to enhance spatial representation and channel-wise adaptability.  
Each CAResBlock includes two $3\times3$ convolutional layers with ReLU activation, followed by an adaptive global average pooling layer.  
The pooled channel descriptor is further processed by a 1D convolution to model inter-channel correlations.  
Finally, the output feature of each CAResBlock is normalized by a LayerNorm2d layer to stabilize the feature distribution during training.  
During this stage, only the parameters of the Alignment Layer are trainable, while all parameters of the SAM encoder and decoder remain frozen.\\
\textbf{Continual Learning Settings.}  
In the continual learning experiments, we sequentially train and evaluate the model across nine medical segmentation tasks:  
ACDC $\rightarrow$ EBHI-SEG $\rightarrow$ 56Nx $\rightarrow$ DN $\rightarrow$ Polyp $\rightarrow$ MSD\_Prostate $\rightarrow$ MSD\_Spleen $\rightarrow$ Promise12 $\rightarrow$ STS-2D.  
Each task is assigned an independent Alignment Layer trained under the same configuration as described above. 
A VAE Router is introduced to achieve automatic task identification and dynamic adapter switching.
Each task-specific VAE consists of a two-layer MLP encoder and decoder, with a latent dimension of 64 and a KL-divergence weight $\beta$ of 16.5.  
The task-specific VAEs are trained for 10 epochs using a learning rate of $5\times10^{-4}$.  
For the fallback mechanism, we perform $K$-fold calibration ($K=5$) to estimate the task-specific threshold $\tau_t$, where $\tau_t$ is chosen as the $p_{97}$ percentile of the ELBO distribution.  
The detailed numerical thresholds and additional implementation details for all comparison methods are provided in the Appendix.

\newcommand{\dropraise}{-0.15ex}
\newcommand{\dropsize}{7pt}
\newcommand{\drop}[2]{%
  #1\kern0.15em%
  \raisebox{\dropraise}[0pt][0pt]{\smash{\fontsize{\dropsize}{\dropsize}\selectfont(#2)}}%
}

\begin{table*}[h]
\centering
\small
\setlength{\tabcolsep}{2.5pt}
\resizebox{\textwidth}{!}{%
\begin{tabular}{l||cc|cc|cc|cc|cc}
\toprule
\multirow{2}{*}{\textbf{Method}} &
\multicolumn{2}{c|}{\textbf{DDTI}} &
\multicolumn{2}{c|}{\textbf{BUSI\_benign}} &
\multicolumn{2}{c|}{\textbf{BUSI\_malignant}} &
\multicolumn{2}{c|}{\textbf{Brain\_Tumor}} &
\multicolumn{2}{c}{\textbf{CAMO}} \\
\cmidrule(lr){2-3}\cmidrule(lr){4-5}\cmidrule(lr){6-7}\cmidrule(lr){8-9}\cmidrule(lr){10-11}
& IoU & BIoU & IoU & BIoU & IoU & BIoU & IoU & BIoU & IoU & BIoU \\
\midrule
\rowcolor{gray!20}
SAM~\cite{kirillov2023segment}
  & 78.08 & 45.98
  & 75.73 & 53.30
  & 69.90 & 33.70
  & 52.85 & 60.89
  & 67.60 & 43.05 \\
MoDA~\cite{yang2024continual}
  & \drop{29.66}{\textcolor{green!60!black}{48.42}} & \drop{20.18}{\textcolor{green!60!black}{25.80}}
  & \drop{10.06}{\textcolor{green!60!black}{65.67}} & \drop{11.45}{\textcolor{green!60!black}{41.85}}
  & \drop{6.24}{\textcolor{green!60!black}{63.66}}  & \drop{6.50}{\textcolor{green!60!black}{27.20}}
  & \drop{24.47}{\textcolor{green!60!black}{28.38}} & \drop{30.59}{\textcolor{green!60!black}{30.30}}
  & \drop{21.51}{\textcolor{green!60!black}{46.09}} & \drop{13.85}{\textcolor{green!60!black}{29.20}} \\
LwF~\cite{li2017learning}
  & \drop{61.34}{\textcolor{green!60!black}{16.74}} & \drop{21.22}{\textcolor{green!60!black}{24.76}}
  & \drop{59.16}{\textcolor{green!60!black}{16.57}} & \drop{37.64}{\textcolor{green!60!black}{15.66}}
  & \drop{30.38}{\textcolor{green!60!black}{39.52}} & \drop{13.07}{\textcolor{green!60!black}{20.63}}
  & \drop{34.20}{\textcolor{green!60!black}{18.65}} & \drop{27.46}{\textcolor{green!60!black}{33.43}}
  & \drop{21.14}{\textcolor{green!60!black}{46.46}} & \drop{12.79}{\textcolor{green!60!black}{30.26}} \\
\rowcolor{gray!20}
CA-SAM
  & \drop{77.62}{\textcolor{green!60!black}{\textbf{0.46}}}  & \drop{45.73}{\textcolor{green!60!black}{\textbf{0.25}}}
  & \drop{72.99}{\textcolor{green!60!black}{\textbf{2.74}}}  & \drop{51.67}{\textcolor{green!60!black}{\textbf{1.63}}}
  & \drop{65.86}{\textcolor{green!60!black}{\textbf{4.04}}}  & \drop{31.66}{\textcolor{green!60!black}{\textbf{2.04}}}
  & \drop{52.85}{\textcolor{red}{\textbf{0.00}}}            & \drop{60.89}{\textcolor{red}{\textbf{0.00}}}
  & \drop{62.37}{\textcolor{green!60!black}{\textbf{5.23}}} & \drop{39.60}{\textcolor{green!60!black}{\textbf{3.45}}} \\
\bottomrule
\end{tabular}%
}
\caption{Zero-shot results of Continual Learning methods. \textcolor{green!60!black}{\textbf{Green}} and \textcolor{red}{\textbf{red}} numbers denote the decrease and increase relative to SAM.
}
\label{tab:zero-shot_on_OOD}
\end{table*}
\vspace{-5pt}

\subsection{Single-dataset Versatility Analysis}
\Cref{tab:Single_Med-Dataset_Versatility_Results} reports results on nine medical image datasets trained and evaluated independently. For each sub-dataset, we configure a dedicated Alignment Layer, achieving the highest IoU and BIoU on the weighted average across all nine. Detailed results for every single-dataset are provided in the Appendix.
Specifically, comparing to Tuning Decoder, HQ\mbox{-}SAM~\cite{sam_hq}, and SAMMed~2D~\cite{cheng2023sammed2d}, our method improves the \textbf{Average IoU} by \(9.75\%\), \(7.24\%\), and \(4.98\%\), respectively; and the \textbf{Average BIoU} by \(12.66\%\), \(8.11\%\), and \(7.55\%\), respectively.
Detailed Table in Appendix also lists the number of trainable parameters for each method. Our approach attains the best overall performance with substantially fewer trainable parameters.

\begin{table}[h]
\centering
\resizebox{0.48\textwidth}{!}{%
\begin{tabular}{lcccc}
\toprule
\textbf{Methods} & \textbf{Parameters} & \textbf{GFLOPs} & \textbf{Avg-IoU} & \textbf{Avg-BIoU} \\
\midrule
SAM~\cite{kirillov2023segment} & --    & --   & 55.08 & 37.67 \\
Tuning Decoder~\cite{kirillov2023segment} & 4.06M  & 669.8   & 70.40 & 53.86 \\
HQ-SAM~\cite{sam_hq}                  & 5.14M  & 678.9 & 72.91 & 58.41 \\
SAMMed 2D~\cite{cheng2023sammed2d}   & 13.31M & 728.2 & \underline{75.17} & \underline{58.97} \\
\rowcolor{gray!20} 
CA-SAM (Ours)                         & \textbf{3.54M} & \textbf{514.3} & \textbf{80.15} & \textbf{66.52} \\
\bottomrule
\end{tabular}%
}
\caption{Single-dataset Versatility Results.}
\label{tab:Single_Med-Dataset_Versatility_Results}
\end{table}

\subsection{Cross-dataset Stability Details}
\noindent\textbf{Quantitative Evaluation.}  
\Cref{tab:continual-learning-results-med} presents the quantitative results under our experimental setup. The key observations are summarized as follows:  
(1) Naive adapter fine-tuning suffers from severe forgetting. Its performance is far below the joint-training upper bound. This indicates that naive adapter updates cannot resist cross-task distributional shifts.  
(2) Classical continual learning methods such as LwF, EWC, ER, and DER moderately mitigate forgetting, demonstrating the necessity of introducing continual learning mechanisms. However, their effectiveness remains limited under pixel-level medical image segmentation with strong distributional shifts.  
(3) Despite their parameter efficiency, the prompt-based method shows limited gains.  
(4) Parameter-fusion methods show limitations when applied to multi-task learning involving cross-modality and cross-organ data.  
(5) MoDA greatly outperforms other CL methods. However, CA-SAM further achieves the highest performance under the same frozen SAM setting, reducing forgetting to nearly zero. Compared with the second-best approach, CA-SAM improves \textit{Last-IoU} and \textit{Last-BIoU} by 5.48\% and 4.43\%, respectively. Notably, methods that use routing exhibit strong robustness (see Appendix) and are inherently invariant to order, whereas methods such as EWC and LwF tend to suffer from catastrophic forgetting when consecutive tasks differ substantially in modality.

\begin{figure*}[!t]  
  \centering  
  \includegraphics[width=0.9\linewidth]{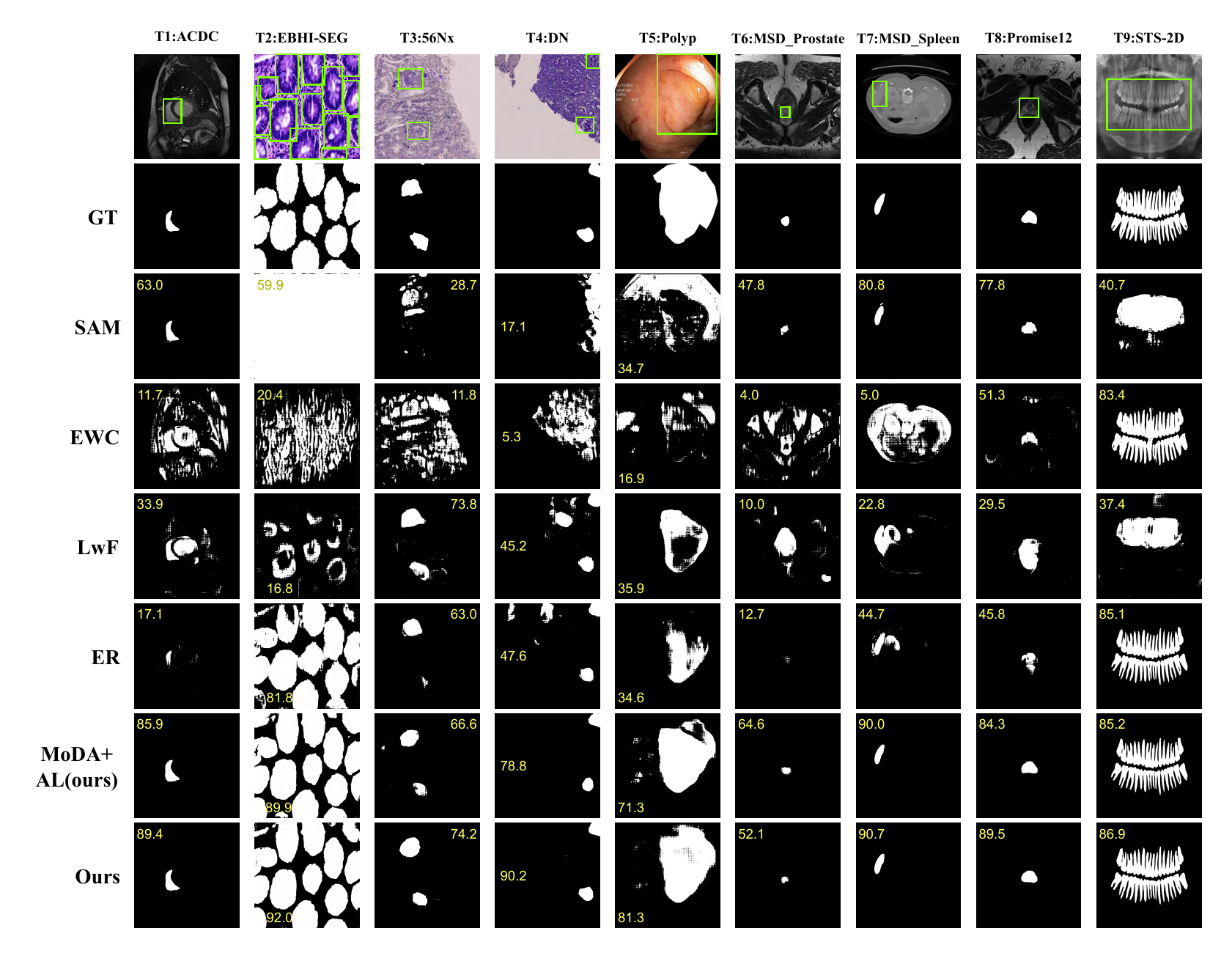}  
  \caption{Qualitative comparison of segmentation results and corresponding IoU scores after continual training on all nine tasks. The figure illustrates the performance differences among competing continual learning methods. AL denotes the proposed Alignment Layer module.}
  \label{fig:visual_results}
\end{figure*}

\noindent\textbf{Visual Evaluation.}
\Cref{fig:visual_results} presents the visual results and IoU scores of different methods after completing continual training on all nine tasks. 
Continual learning methods such as ER, EWC, and LwF still suffer from forgetting, and their performance even falls below SAM on several datasets.In contrast, our method delivers accurate and stable segmentation across all tasks, showing significant improvements in accuracy and stability over SAM. Even on challenging tasks (e.g., T2 and T4), where other methods fail and SAM produces incorrect masks, our approach still produces high-quality and coherent segmentation results.

\subsection{Evaluation on Out-of-Distribution Data}
To assess whether continual learning compromises the generalization ability of SAM, we evaluate each model’s zero-shot segmentation performance on unseen domains after completing training on all medical tasks. 
All comparison methods are reproduced using their publicly available implementations, and the results are reported in \Cref{tab:zero-shot_on_OOD}. 
Due to the lack of effective mechanisms for preserving generalization, models trained with LwF and MoDA suffer from a severe loss of SAM’s zero-shot capability. 
In particular, MoDA’s hard routing often misassigns OOD samples to incorrect adapters, causing feature mismatch and eroding the base model’s general knowledge.
In contrast, our proposed \textit{CA-SAM} introduces a VAE-based task discrimination and OOD fallback mechanism that maximally preserves SAM’s inherent generalization. 
Experimental results on five unseen datasets demonstrate that the VAE Router achieves a weighted average discrimination accuracy of 99.42\% (DDTI: 99.37\%, BUSI\_benign: 96.34\%, BUSI\_malignant: 92.38\%, Brain\_Tumor: 100.00\%, CAMO: 92.00\%).
Most OOD samples are correctly routed back to SAM for zero-shot inference, enabling our method to achieve zero-shot performance on unseen domains that is nearly equivalent to the SAM upper bound.


\begin{table}[t]
\centering
\renewcommand{\arraystretch}{0.9}
\begin{minipage}{0.45\textwidth} 
  \centering
  \begin{tabular}{ccc}
  \toprule
  \textbf{56Nx} & \textbf{TV Distance} \(\downarrow\) & \textbf{JS Divergence} \(\downarrow\) \\ 
  \midrule
  Before vs Adapter & 0.2830 & 0.2398 \\ 
  \rowcolor{gray!20} 
  After vs Adapter  & 0.2742 & 0.2351 \\ 
  \bottomrule
  \end{tabular}
\end{minipage}

\vspace{8pt}

\begin{minipage}{0.45\textwidth}
  \centering
  \begin{tabular}{ccc}
  \toprule
  \textbf{DN} & \textbf{TV Distance} \(\downarrow\) & \textbf{JS Divergence} \(\downarrow\) \\ 
  \midrule
  Before vs Adapter & 0.2840 & 0.2404 \\ 
  \rowcolor{gray!20} 
  After vs Adapter  & 0.2832 & 0.2365 \\ 
  \bottomrule
  \end{tabular}
\end{minipage}

\caption{Feature Distribution Distance Comparison in 56Nx, DN}
\label{tab:TV_JS}

\end{table}

\subsection{Explainability}
To verify whether the Alignment Layer can align the feature distributions of different datasets to their respective optimal distribution spaces, we conducted experiments using the 56Nx and DN datasets. We compared the feature maps output by the SAM Encoder after Adapter fine-tuning~\cite{wu2025medical} as a baseline, and calculated the distance between the features before and after the Alignment Layer and the baseline features. We selected the TV (Total Variation) and JS (Jensen-Shannon) divergence metrics for comparison. See the Appendix for details on the metric formulations. 

\begin{figure}[h]
  \centering
  \begin{minipage}[t]{0.40\linewidth}
    \centering
    \includegraphics[width=\linewidth]{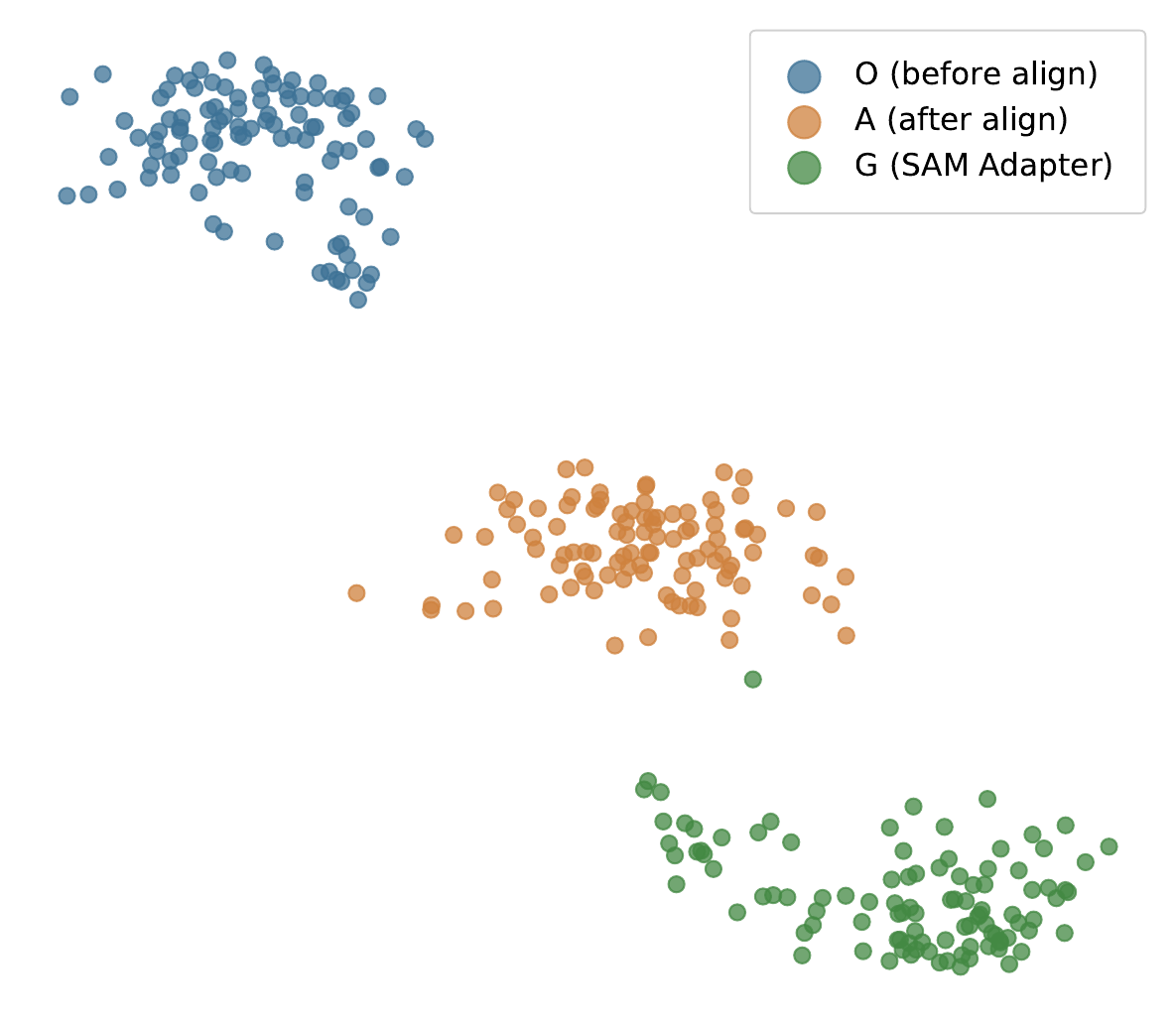}
  \end{minipage}
  \hspace{0.05\linewidth}
  \begin{minipage}[t]{0.40\linewidth}
    \centering
    \includegraphics[width=\linewidth]{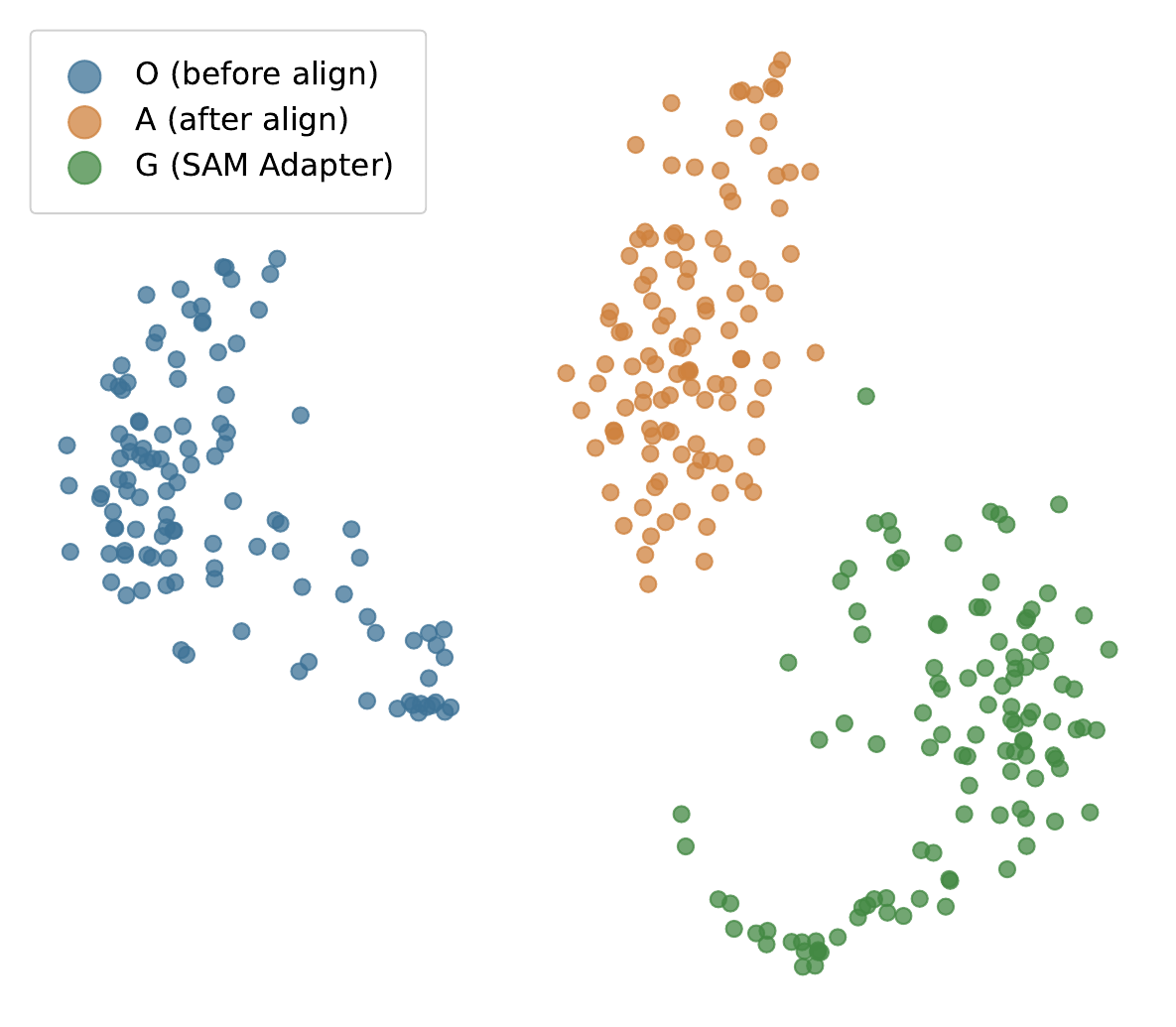}
  \end{minipage}

  \caption{Two TSNE visualization comparison images: the left image is 56Nx dataset, and the right image is DN dataset.}
  \label{fig:two_TSNE_horizontal}
\vspace{-10pt}
\end{figure}

The results are shown in the \Cref{tab:TV_JS} and the \Cref{fig:two_TSNE_horizontal}.
The experiments show that the feature maps after the Alignment Layer are closer to the outputs of the SAM Encoder tuning than those before the Alignment Layer, confirming that the Alignment Layer effectively aligns the features within the target medical domain.

\begin{figure}[h]  
  \centering
  \includegraphics[width=\linewidth]{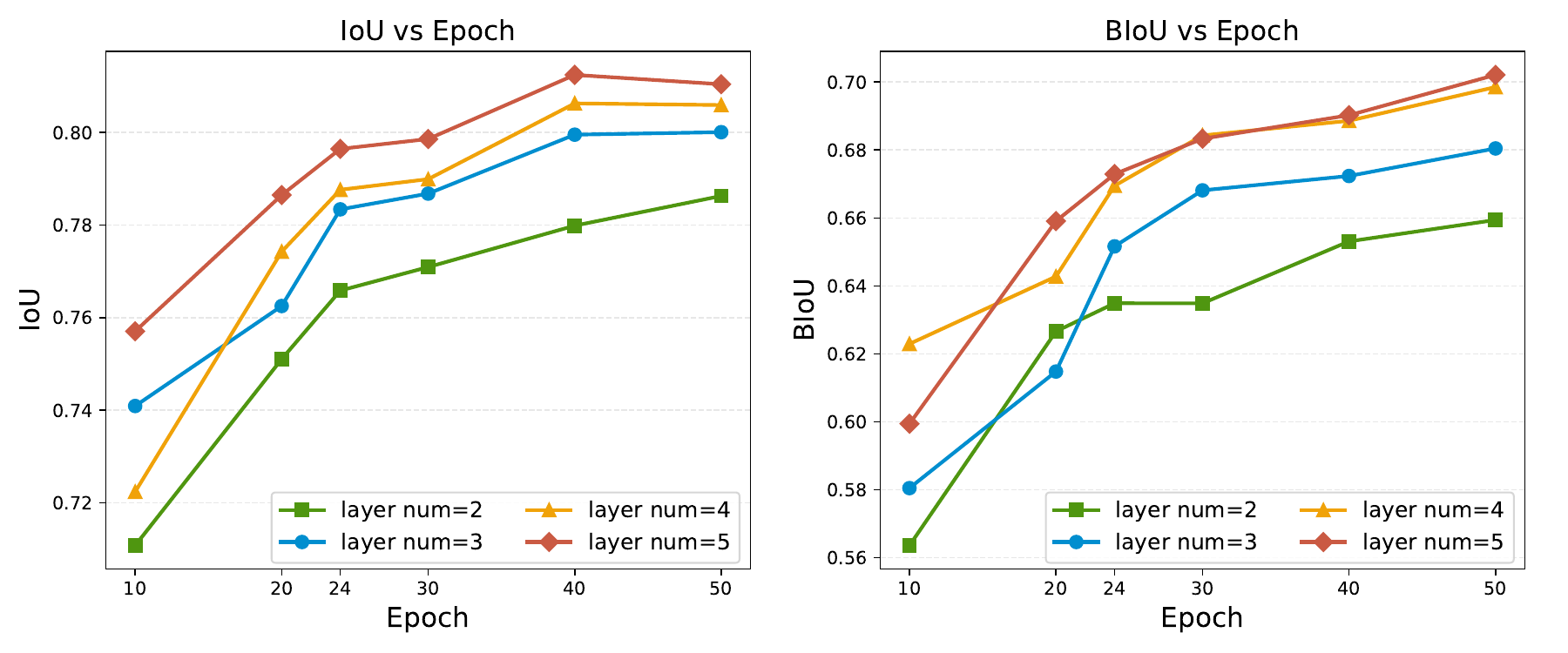}  
  \caption{Ablation Study with Number of Alignment Layers.}
  \label{fig:iou_epoch_vis}

\end{figure}

\begin{table*}[t]
\centering
\resizebox{\textwidth}{!}{%
\begin{tabular}{l||c| ccc| ccc}
\toprule\toprule
\multirow{2}{*}{\textbf{Method}} & \multirow{2}{*}{\textbf{Hyper-Param}} & \multicolumn{3}{c|}{\textbf{IoU on Med}} & \multicolumn{3}{c}{\textbf{BIoU on Med}} \\
\cmidrule(lr){3-5} \cmidrule(lr){6-8}
 & & \textbf{Last-IoU} & \textbf{Avg-IoU} & \textbf{FF-IoU} & \textbf{Last-BIoU} & \textbf{Avg-BIoU} & \textbf{FF-BIoU} \\
\midrule
Global Average Pooling & $\beta = 16.5$ & 75.93 & 76.87 & 1.65 & 59.63 & 59.38 & 0.42 \\
Mean                   & $\beta = 16.5$ & 75.86 & 76.91 & 1.61 & 59.72 & 59.46 & 0.38 \\
Flatten                & $\beta = 16.5$ & 24.06 & 33.64 & 7.55 & 20.95 & 28.94 & 6.34 \\
CLS Token              & $\beta = 16.5$ & 49.71 & 48.19 & 0.01 & 31.23 & 28.42 & 0.02 \\
Attn Pooling (With Params) & $\beta = 16.5$, temp = 1 & 71.03 & 72.59 & 1.14 & 56.39 & 56.52 & 0.71 \\
\rowcolor{gray!20}
Attn Pooling (No Params)   & $\beta = 16.5$, temp = 1 & \textbf{76.41} & \textbf{77.07} & 1.28 & \textbf{59.91} & \textbf{59.51} & 0.27 \\
\bottomrule\bottomrule
\end{tabular}%
}
\caption{Comparison of different pooling methods and their performance metrics.}
\label{tab:Comparison of different pooling}
\end{table*}


\subsection{Ablation Study}

\subsubsection{Single-dataset Versatility Ablation}

We experimented with different numbers of Blocks in Alignment Layer, and collected the average IoU/BIoU across different epochs. The experimental results are shown in the \Cref{fig:iou_epoch_vis}.  We found that, increasing the number of Alignment Layers can enhance its fitting ability, thus correspondingly improving the performance. The results demonstrate the effectiveness of proposed CA-SAM.

\subsubsection{Cross-dataset Stability Ablation}
\noindent\textbf{Feature Pooling Ablation.}
We evaluate multiple feature-pooling designs and report IoU/BIoU after sequential training and testing on nine medical datasets, as summarized in the \Cref{tab:Comparison of different pooling}. The candidates include GAP (global average pooling over spatial dimensions of the encoder feature map), Mean pooling, Flatten (reshape a 2D feature map into a vector followed by a linear projection to the target dimension), CLS token (take the Transformer’s [CLS] output directly), parameter-free attention pooling (compute the \(\ell_2\) norm per spatial location on the feature map, normalize with Softmax to obtain attention over \((H,W)\), and use it to weighted-sum the original \((B,C,H,W)\) features into \((B,C)\)), and learnable attention pooling (as above, but with trainable parameters to generate attention weights).
Across all settings, attention pooling yields the strongest segmentation performance, consistently outperforming the alternatives by better focusing on salient regions and providing the VAE with more discriminative global features. Notably, the parameter-free attention pooling achieves the best task separability without introducing any additional parameters, delivering the overall best results among the tested strategies.

\noindent\textbf{Router Threshold ($\tau$) Ablation.}  
The router threshold \( \tau \) is the key hyperparameter for rejecting unknown tasks. To improve decision stability, we calibrate a separate threshold $\tau_t$ for each task $t$. For this calibration, we perform K-fold cross-validation on the training set using the task’s VAE $V_t$. We evaluate several rules for setting $\tau_t$ based on the ELBO score distribution on the training set, namely $\mu+2\sigma$ and the $p_{95}$/$p_{97}$/$p_{99}$ percentiles.The results in \Cref{tab:vae_tau_comparison} reveal a trade-off between \textit{known-task routing accuracy} and \textit{zero-shot (OOD) performance}.
 A lenient threshold such as $p_{99}$ biases the router toward “known,” which improves seen-task continual segmentation (IoU on Med 76.04\%) but degrades OOD discrimination (Acc 98.76\%). In contrast, $p_{97}$ applies a tighter cutoff, yielding stronger OOD accuracy (Acc 99.42\%) \textit{without material loss} on seen-task continual segmentation. We therefore adopt $p_{97}$ as our calibration rule, striking a better balance between OOD detection and segmentation on known tasks.

\begin{table}[h]
\centering
\setlength{\tabcolsep}{2.8pt}
\renewcommand{\arraystretch}{0.95}
\small
\begin{tabular}{c|ccccc}
\toprule
\textbf{Parameter} & \textbf{IoU (Med)} & \textbf{BIoU (Med)} & \textbf{IoU} & \textbf{BIoU} & \textbf{Acc} \\
\midrule
$\mu + 2\sigma$ & 75.84 & 59.66 & 55.87 & 58.27 & 99.33 \\
$p_{95}$            & 75.83 & 59.56 & \textbf{56.08} & \textbf{58.39} & \textbf{99.66} \\
$p_{97}$             & 76.02 & 59.64 & 55.92 & 58.29 & 99.42 \\
$p_{99}$            & \textbf{76.04} & \textbf{59.71} & 55.48 & 58.08 & 98.76 \\
\bottomrule
\end{tabular}
\vspace{-3pt}
\caption{
\textbf{Comparison of CA-SAM configurations under different $\tau$ thresholds.}
“Acc” denotes OOD discrimination accuracy.
}
\label{tab:vae_tau_comparison}

\end{table}


\section{Conclusion}
In this paper, we demonstrate that SAM provides strong segmentation priors and considerable potential for continual learning. However, fine-tuning all parameters incurs considerable computational and parameter overheads. To balance performance and efficiency, we introduce a lightweight and plug-and-play Alignment Layer that aligns features between the encoder and decoder, significantly reducing trainable parameters while maintaining high segmentation accuracy. Building on the Alignment Layer, we propose Continual Alignment for SAM (CA-SAM), an exemplar-free continual learning framework that automatically identifies tasks and routes inputs to the appropriate Alignment Layer or the frozen SAM for OOD fallback. By providing robustness to varying task orders and strong performance across diverse medical datasets, CA-SAM highlights a promising direction for enabling scalable and reliable continual learning of pretrained foundation models in \mbox{medical image segmentation}.


{
    \small
    \bibliographystyle{ieeenat_fullname}
    \bibliography{main}
}


\clearpage
\setcounter{page}{1}
\maketitlesupplementary

\setcounter{secnumdepth}{3}
\renewcommand{\thesubsection}{\Alph{subsection}}
\renewcommand{\thesubsubsection}{\thesubsection.\arabic{subsubsection}}

\subsection{Dataset Details}
We selected nine medical datasets, each representing different anatomical regions and imaging modalities. Detailed information about these datasets is provided below.

\textbf{ACDC:} The ACDC dataset consists of 1632 training images and 177 test images. The dataset modality is MR (Magnetic Resonance) imaging. The segmentation targets include three parts of the heart: the left ventricle, myocardium, and right ventricle.

\textbf{EBHI-SEG:} The EBHI-SEG dataset contains 1701 training images and 487 test images. The dataset modality is pathological imaging, with the segmentation target being colon cancer (affected areas).

\textbf{56Nx:} The 56Nx dataset includes 558 training images and 463 test images. The dataset modality is pathological imaging, with the segmentation target being the glomerulus.

\textbf{DN:} The DN dataset contains 724 training images and 391 test images. The dataset modality is pathological imaging, and the segmentation target is the glomerulus.

\textbf{Ployp:} The Ployp dataset consists of 804 training images and 196 test images. The dataset modality is RGB imaging, and the segmentation target is the spleen.

\textbf{MSD\_prostate:} The MSD\_prostate dataset includes 419 training images and 53 test images. The dataset modality is MR T2 (Magnetic Resonance Imaging), with the segmentation target being two regions of the prostate: the peripheral zone and transition zone.

\textbf{MSD\_Spleen:} The MSD\_Spleen dataset contains 876 training images and 146 test images. The dataset modality is CT imaging, and the segmentation target is the spleen.

\textbf{promise12:} The promise12 dataset includes 712 training images and 66 test images. The dataset modality is MR (Magnetic Resonance) imaging, and the segmentation target is the prostate.

\textbf{STS-2D:} The STS-2D dataset consists of 1700 training images and 70 test images. The dataset modality is X-ray imaging, and the segmentation target is the teeth.

\begin{figure}[H]
  \centering
  \includegraphics[width=\linewidth]{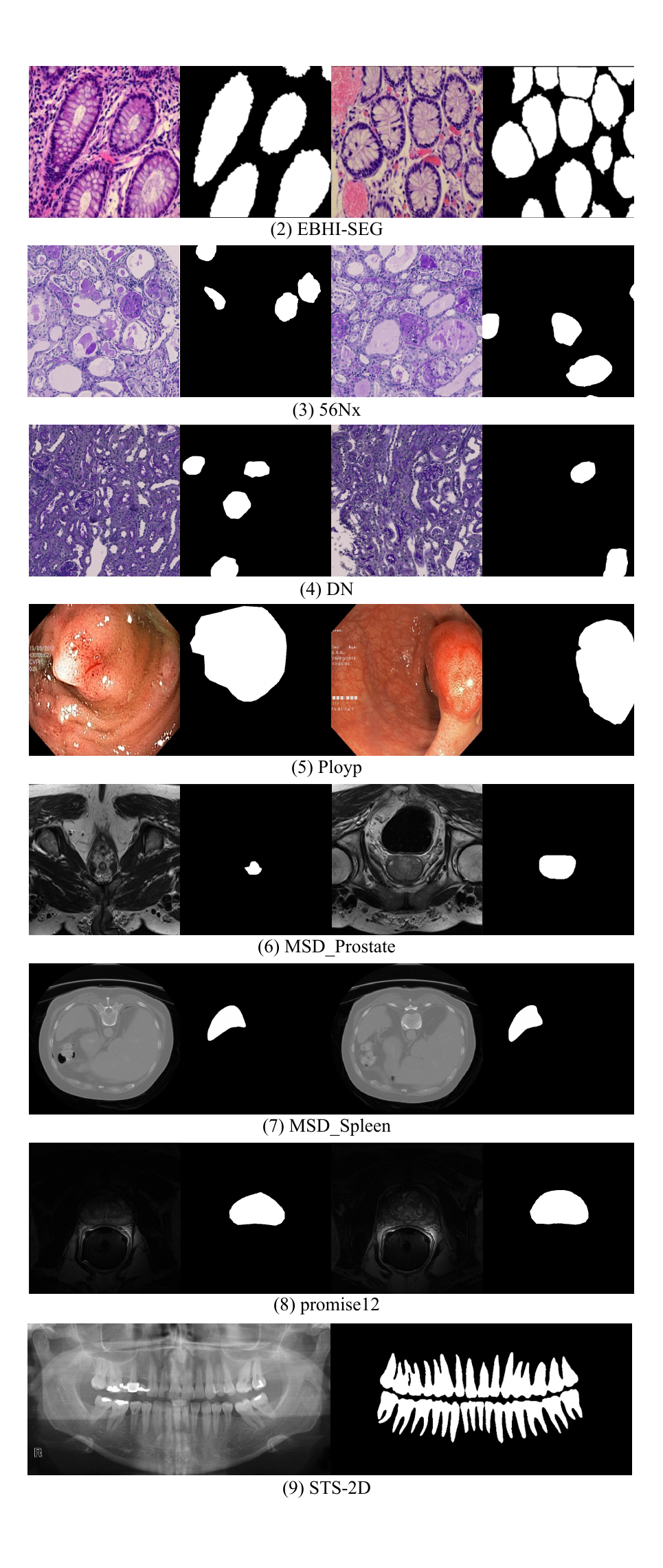} 
  \vspace{-40pt}
  \caption{\textbf{Datasets of Task 1-9 :} (1)-(9) show the examples of the images and corresponding masks for nine medical dataset.}
  \label{fig:dataset_example}
\end{figure}

\Cref{fig:dataset_example} present representative examples of the images and corresponding masks for each dataset.

\begin{figure}[!h]
  \centering
  \includegraphics[width=\linewidth]{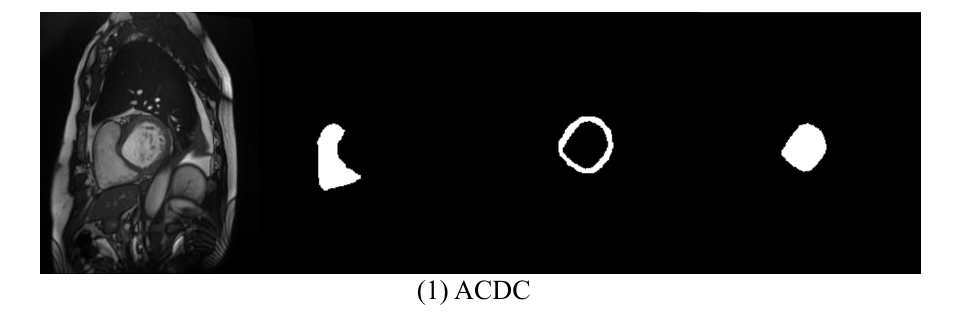}
  \caption*{}
  \vspace{-20pt}
  \label{fig:acdc}
\end{figure}

\subsection{Experiment Evaluation Metrics}

In the Cross-dataset Stability experiment, we selected three metrics: \textbf{Last-IoU}, \textbf{Avg-IoU} and \textbf{FF-IoU} to evaluate the continual segmentation performance of different methods across nine datasets. To evaluate the final overall accuracy, we employ the metrics Last-IoU and Last-BIoU to measure the segmentation performance across all tasks after the completion of all sequential tasks. We define the average performance after training \( t \) tasks, denoted as \( \text{IoU}_t \),
\begin{equation}
\text{IoU}_t = \frac{1}{N_t} \sum_{k=1}^{t} n_k \, \text{IoU}_{k,t}, 
\quad 
\text{BIoU}_t = \frac{1}{N_t} \sum_{k=1}^{t} n_k \, \text{BIoU}_{k,t},
\end{equation}

where \( \text{IoU}_{k,t} \) and \( \text{BIoU}_{k,t} \) represent the weighted IoU/BIoU evaluated on the test set of the \( k \)-th task after training on \( t \) tasks, and \( n_k \) denotes the number of images in the test set of the \( k \)-th task and \( N_t= \sum_{k=1}^{t}n_k\).
Then, the Last-IoU and Last-BIoU are defined as
\begin{equation}
\text{Last-IoU} = \text{IoU}_N, \quad \text{Last-BIoU} = \text{BIoU}_N.
\end{equation}

To illustrate the average segmentation performance throughout the training process during sequential learning, we also use the Avg-IoU and Avg-BIoU metrics, as described below:
\begin{equation}
\text{Avg-IoU} = \frac{1}{N} \sum_{t=1}^{N} \text{IoU}_t, \quad \text{Avg-BIoU} = \frac{1}{N} \sum_{t=1}^{N} \text{BIoU}_t.
\end{equation}

To measure forgetting performance, we use \( f_{k,t} \) as the forgetting on task \( t \) after training on all \( t \) tasks,
\begin{equation}
f_{k,t} = \max_{j \in \{1, \dots, t-1\}} \left( \text{IoU}_{k,j} - \text{IoU}_{k,t} \right).
\end{equation}

Then, the average forgetting measure, defined as FF-IoU, can be computed after training on all 
\emph{N} tasks.
\begin{equation}
\text{FF-IoU} = \frac{1}{N-1} \sum_{k=1}^{N-1} f_{k,N}. 
\end{equation}

The metric \( \text{FF-BIoU} \) is calculated in a similar way.



\subsection{Single-dataset Versatility Analysis}
\label{sec:Appendix-Single-dataset-Versatility-Analysis}

Table~\ref{tab:Detailed-Single-Med-Dataset-Versatility-Results} in the Appendix reports the complete single-dataset experiment results for training and testing the proposed \emph{Alignment Layer} on each of the nine medical datasets. Specifically, our method attains the best performance on five out of nine datasets (ACDC, EBHI-SEG, 56Nx, DN and STS-2D) as well as on the weighted average across datasets. Notably, on \textit{56Nx} and \textit{DN} we surpass the second-best approach by \textbf{10.3\%} and \textbf{24.9\%} in IoU, and by \textbf{9.2\%} and \textbf{28.0\%} in BIoU, respectively. Taken together, these results provide strong evidence for the effectiveness of the proposed Alignment Layer.

\begin{table*}[t]
\centering

\begin{minipage}{\textwidth}
  \centering
  \resizebox{\textwidth}{!}{%
  \begin{tabular}{*{12}{c}}
    \toprule
    \textbf{Methods(Iou)} & \textbf{Parameters} & \textbf{ACDC} & \textbf{EBHI\_SEG} & \textbf{56Nx} & \textbf{DN} & \textbf{Polyp} & \textbf{MSD\_Prostate} & \textbf{MSD\_Spleen} & \textbf{promise12} & \textbf{STS2D} & \textbf{Average Iou} \\
    \midrule
    Zero-Shot~\cite{kirillov2023segment} & 0 M & 61.47 & 63.98 & 29.06 & 32.12 & 66.61 & 60.55 & 81.39 & 83.23 & 65.10 & 55.08 \\
    Tuning Decoder & 4.06M & 68.45 & 89.53 & 43.05 & 50.96 & 81.49 & 67.42 & $\mathbf{91.98}$ & $\mathbf{90.58}$ & 82.38 & 70.40 \\
    HQ-SAM~\cite{sam_hq} & 5.14M & 77.05 & 85.07 & 47.11 & 58.34 & 78.46 & $\mathbf{72.33}$ & 90.27 & 87.65 & 81.88 & 72.91 \\
    SAMMed 2D~\cite{cheng2023sammed2d} & 13.31M & 78.27 & $\mathbf{89.58}$ & 55.29 & 52.20 & $\mathbf{86.87}$ & 71.46 & 91.72 & 86.69 & 82.16 & 75.17 \\
    CA-SAM(Ours) & $\mathbf{3.54M}$ & $\mathbf{80.75}$ & 89.14 & $\mathbf{65.92}$ & $\mathbf{83.23}$ & 65.02 & 62.71 & 86.18 & 84.52 & $\mathbf{86.03}$ & $\mathbf{80.15}$ \\
    \bottomrule
  \end{tabular}%
  }
\end{minipage}

\vspace{3pt}

\begin{minipage}{\textwidth}
  \centering
  \resizebox{\textwidth}{!}{%
  \begin{tabular}{*{12}{c}}
    \toprule
    \textbf{Methods(BIou)} & \textbf{Parameters} & \textbf{ACDC} & \textbf{EBHI\_SEG} & \textbf{56Nx} & \textbf{DN} & \textbf{Polyp} & \textbf{MSD\_Prostate} & \textbf{MSD\_Spleen} & \textbf{promise12} & \textbf{STS2D} & \textbf{Average BIou} \\
    \midrule
    Zero-Shot~\cite{kirillov2023segment}       & 0 M     & 57.83 & 19.85 & 22.17 & 22.63 & 42.48 & 53.11 & 75.44 & 64.42 & 36.45 & 37.67 \\
    Tuning Decoder  & 4.06M & 65.44 & 54.09 & 22.28 & 29.41 & 55.52 & 64.55 & 88.00 & $\mathbf{77.56}$ & 80.85 & 53.86 \\
    HQ-SAM~\cite{sam_hq}          & 5.14M     & 73.18 & 50.18 & 33.46 & 43.04 & 56.06 & $\mathbf{65.24}$ & 86.52 & 71.97 & 79.63 & 58.41 \\
    SAMMed 2D~\cite{cheng2023sammed2d}       & 13.31M & 73.09 & 54.44 & 35.99 & 31.71 & $\mathbf{65.65}$ & 64.88 & $\mathbf{89.22}$ & 73.93 & 80.84 & 58.97 \\
    CA-SAM(Ours)   & $\mathbf{3.54M}$ & $\mathbf{75.90}$ & $\mathbf{62.93}$ & $\mathbf{45.18}$ & $\mathbf{71.01}$ & 42.16 & 53.62 & 87.19 & 69.52 & $\mathbf{84.75}$ & $\mathbf{66.52}$ \\
    \bottomrule
  \end{tabular}%
  }
\end{minipage}
\caption{Detailed Single Med-Dataset Versatility Results}
\label{tab:Detailed-Single-Med-Dataset-Versatility-Results}
\end{table*}

\subsection{Cross-dataset Stability Details}
\subsubsection{Pipeline of Algorithms}
To assess the capability of contrastive methods for medical image segmentation in continual learning, we adapt and modify each method for the alignment-based SAM framework as follows:\\
1) For \textbf{LwF}, the alignment layers trained on the previous task act as the teacher network, and a knowledge distillation loss constrains the current alignment layers to retain consistent outputs with the teacher while learning new knowledge, thereby mitigating catastrophic forgetting. \\
2) For \textbf{EWC}, we estimate the importance of each parameter using the Fisher Information Matrix and penalize changes to crucial parameters during new task training to preserve previously learned knowledge. \\
3) For \textbf{ER}, a memory bank stores a small set of samples from past tasks, which are replayed and jointly trained with new task data to maintain prior performance. \\
4) For \textbf{DER}, this method further introduces knowledge distillation on the basis of ER. It constrains the consistency between the current model and historical model predictions by storing and replaying past samples in memory and matching the network's Logits sampled throughout the optimization trajectory.\\
5)For \textbf{L2P}, we maintain a cumulative prompt pool that uses SAM image embeddings as query keys to retrieve the most relevant prompts, while a slot-based allocation mechanism ensures task-wise isolation and efficient prompt utilization.\\
6)For \textbf{MoDA}, we introduce a task classifier by augmenting the encoder with a [CLS] token that captures global task information; during inference, the classifier automatically routes the input image to its corresponding historical alignment layer according to the [CLS] token.\\
7)For \textbf{EMR}, we employ a parameter space merging strategy on the alignment layer. This method defines the task vectors $\tau_i$ as the weight increments from each historical task. These vectors are then aggregated into a single unified task vector $\tau_{\text{uni}}$ via an Electing procedure, which selects the parameter with the largest magnitude that is consistent in sign across all tasks. During inference, $\tau_{\text{uni}}$ is subject to task-specific modulation: an alignment mask ($M_t$) is applied to filter $\tau_{\text{uni}}$ by zeroing out parameters that conflict in direction with the specific task vector, and a rescaling factor $\lambda_t$ is used to calibrate the magnitude of the modulated vector to match the original task vector's scale, thereby allowing the single layer to efficiently store and recall knowledge for each specific task.\\
8)We further include \textbf{Joint Training} as an upper bound, where all task datasets are trained simultaneously. \\
9)\textbf{Naive approach} as a lower bound. This method is a simple Sequential Fine-Tuning baseline under the Frozen SAM framework, where only the alignment layers are sequentially updated without any CL mechanism to quantify cross-task forgetting.
\begin{table}[h]
\centering
\small
\renewcommand{\arraystretch}{0.95}
\setlength{\tabcolsep}{2pt}

\begin{tabular}{p{0.04\linewidth} p{0.90\linewidth}}
\toprule
\multicolumn{2}{l}{\textbf{Algorithm 1}. Continual Alignment for SAM with LwF} \\
\midrule
\multicolumn{2}{l}{\textbf{Input:} Pre-trained SAM $\theta$, Tasks $\mathcal{D}=\{D^{tr}_1,\ldots,D^{tr}_N\}$} \\
\multicolumn{2}{l}{\textbf{Output:} $\mathcal{A}_N$} \\

1: & Initialize $\mathcal{A}_1$ and train on $D^{tr}_1$. \vspace{-3pt} \\

2: & \textbf{for} $t=2,\ldots,N$ \textbf{do} \\
3: & \quad Set teacher $\mathcal{A}_{t-1}$ and initialize $\mathcal{A}_t \leftarrow \mathcal{A}_{t-1}$. \\
4: & \quad \textbf{for} $(x,y) \in D^{tr}_t$ \textbf{do} \\
5: & \quad\quad $\hat{m}_t = f_\theta(x; \mathcal{A}_t)$, $\hat{m}_{t-1} = f_\theta(x; \mathcal{A}_{t-1})$. \\
6: & \quad\quad Update $\mathcal{A}_t$ with $\mathcal{L} = \mathcal{L}_{align}(\hat{m}_t, y) + \lambda \mathcal{L}_{mask}(\hat{m}_t, \hat{m}_{t-1})$. \\
7: & \quad \textbf{end for} \\
8: & \textbf{end for} \\
9: & \textbf{return} $\mathcal{A}_N$ \\
\bottomrule
\end{tabular}
\end{table}

\begin{table}[h]
\centering
\small
\renewcommand{\arraystretch}{0.95}
\setlength{\tabcolsep}{2pt}

\begin{tabular}{p{0.04\linewidth} p{0.90\linewidth}}
\toprule
\multicolumn{2}{l}{\textbf{Algorithm 2}. Continual Alignment for SAM with EWC} \\
\midrule
\multicolumn{2}{l}{\textbf{Input:} Pre-trained SAM $\theta$, Tasks $\mathcal{D}=\{D^{tr}_1,\ldots,D^{tr}_N\}$} \\
\multicolumn{2}{l}{\textbf{Output:} $\mathcal{A}_N$, $\mathcal{F}_N$} \\

1: & Train $\mathcal{A}_1$ on $D^{tr}_1$ and compute Fisher $\mathcal{F}_1$. \vspace{-3pt} \\

2: & \textbf{for} $t=2,\ldots,N$ \textbf{do} \\
3: & \quad Initialize $\mathcal{A}_t \leftarrow \mathcal{A}_{t-1}$. \\
4: & \quad \textbf{for} $(x,y) \in D^{tr}_t$ \textbf{do} \\
5: & \quad\quad Compute $\mathcal{L}_{align}$ on $D^{tr}_t$\\
6: & \quad\quad Compute $\mathcal{L}_{ewc}=\sum_i\mathcal{F}_{t-1,i}(\mathcal{A}_{t,i}-\mathcal{A}_{t-1,i})^2$. \\
7: & \quad\quad Update $\mathcal{A}_t$ using 
      $\mathcal{L} = \mathcal{L}_{align} + \lambda \mathcal{L}_{ewc}$. \\
8: & \quad \textbf{end for} \\
9: & \quad Estimate new $\mathcal{F}_t$.\\
10: & \textbf{end for} \\
11: & \textbf{return} $\mathcal{A}_N$, $\mathcal{F}_N$ \\
\bottomrule
\end{tabular}
\end{table}


\begin{table}[!h]
\centering
\small
\renewcommand{\arraystretch}{0.95}
\setlength{\tabcolsep}{2pt}

\begin{tabular}{p{0.05\linewidth} p{0.89\linewidth}}
\toprule
\multicolumn{2}{l}{\textbf{Algorithm 3}. Continual Alignment for SAM with ER} \\
\midrule
\multicolumn{2}{p{0.94\linewidth}}{\textbf{Input:} Pre-trained SAM $\theta$, Tasks $\mathcal{D}=\{D^{tr}_1,\ldots,D^{tr}_N\}$} \\
\multicolumn{2}{p{0.94\linewidth}}{\textbf{Output:} $\mathcal{A}_N$, $\mathcal{M}$}\\
1:  & Initialize Memory Bank $\mathcal{M} \leftarrow \emptyset$. \\
2:  & \textbf{for} $t=1,\ldots,N$ \textbf{do} \\
3:  & \quad initialize $\mathcal{A}_t \leftarrow \mathcal{A}_{t-1}$ \\
4:  & \quad \textbf{for} $(x, y) \in D^{tr}_t \cup \mathcal{M}$ \textbf{do} \\
5:  & \quad\quad Compute $\mathcal{L}_{align}$ on $D^{tr}_t$\\
6:  & \quad\quad Update $\mathcal{A}_t$ with loss $\mathcal{L}_{align}$. \\
7:  & \quad \textbf{end for} \\
8:  & \quad Select exemplars from $D^{tr}_t$ to $\mathcal{M}$. \\
9: & \textbf{end for} \\
10: & \textbf{return} $\mathcal{A}_N$, $\mathcal{M}$ \\
\bottomrule
\end{tabular}
\end{table}


\begin{table}[!h]
\centering
\small
\renewcommand{\arraystretch}{0.95}
\setlength{\tabcolsep}{2pt}

\begin{tabular}{p{0.05\linewidth} p{0.89\linewidth}}
\toprule
\multicolumn{2}{l}{\textbf{Algorithm 4}. Continual Alignment for SAM with DER} \\
\midrule
\multicolumn{2}{p{0.94\linewidth}}{\textbf{Input:} Pre-trained SAM $\theta$, Tasks $\mathcal{D}=\{D^{tr}_1,\ldots,D^{tr}_N\}$}\\
\multicolumn{2}{p{0.94\linewidth}}{\textbf{Output:} $\mathcal{A}_N$, $\mathcal{M}$}\\
1: & Initialize Memory Bank $\mathcal{M} \leftarrow \emptyset$. \\
2: & \textbf{for} $t=1,\ldots,N$ \textbf{do} \\
3: & \quad initialize $\mathcal{A}_t \leftarrow \mathcal{A}_{t-1}$ \\
4: & \quad \textbf{for} $(x, y) \in D^{tr}_t$ \textbf{do} \\
5: & \quad\quad $(x^\prime, z^\prime) \leftarrow \text{Sample}(\mathcal{M})$ \\
6: & \quad\quad $Z_t = E_\theta(x)$; $\tilde{Z}_t = \mathcal{A}_t(Z_t)$; $h_t = D_\theta(\tilde{Z}_t)$ \\
7: & \quad\quad $\mathcal{L}_{new} = \mathcal{L}_{align}(y, h_t)$ \quad \\
8: & \quad\quad $Z' = E_\theta(x')$; $\tilde{Z}^\prime = \mathcal{A}_t(Z')$; $h^\prime= D_\theta(\tilde{Z}')$ \\
9: & \quad\quad $\mathcal{L}_{distill} = ||z' - h'||_2^2$ \quad \\
10: & \quad\quad $\mathcal{L}_{total} = \mathcal{L}_{new} + \alpha \mathcal{L}_{distill}$ \\
11: & \quad\quad Update $\mathcal{A}_t$ with loss $\mathcal{L}_{total}$. \\
12: & \quad\quad $\mathcal{M} \leftarrow \text{ReservoirUpdate}(\mathcal{M}, (x, h_t))$ \quad  \\
13: & \quad \textbf{end for} \\
14: & \textbf{end for} \\
15: & \textbf{return} $\mathcal{A}_N$, $\mathcal{M}$ \\
\bottomrule
\end{tabular}
\end{table}

\begin{table}[!h]
\centering
\small
\renewcommand{\arraystretch}{0.95}
\setlength{\tabcolsep}{2pt}

\begin{tabular}{p{0.04\linewidth} p{0.90\linewidth}}
\toprule
\multicolumn{2}{l}{\textbf{Algorithm 5}. Continual Alignment for SAM with L2P} \\
\midrule
\multicolumn{2}{l}{\textbf{Input:} Pre-trained SAM $\theta$, Pre-trained Alignment layer $\mathcal{A}$} \\
\multicolumn{2}{l}{Tasks $\mathcal{D}=\{D^{tr}_1,\ldots,D^{tr}_N\}$} \\
\multicolumn{2}{l}{\textbf{Output:} Prompt Pool $\mathcal{P}_N$} \\

1: & Initialize Prompt Pool $\mathcal{P}$. \\
2: & If using task slots: assign each task its prompt range. \vspace{-3pt} \\

3: & \textbf{for} $t=1,\ldots,N$ \textbf{do} \\
4: & \quad \textbf{for} $(x,y) \in D^{tr}_t$ \textbf{do} \\
5: & \quad\quad $Z = E_\theta(x)$; $\tilde{Z} = \mathcal{A}_t(Z)$. \\
6: & \quad\quad retrieve top-$k$ prompts $p_k$ from $\mathcal{P}$. \\
6: & \quad\quad  $\hat{m} = D_\theta(\tilde{Z}; p_k)$. \\
7: & \quad\quad Compute $\mathcal{L}_{align}(\hat{m}, y)$ and $\mathcal{L}_{key-match}$. \\
8: & \quad\quad Update $\mathcal{P}$. \\
9: & \quad \textbf{end for} \\
10: & \quad Save $\mathcal{P}_t$ for next task. \\
11: & \textbf{end for} \\
12: & \textbf{return} $\mathcal{P}_N$ \\
\bottomrule
\end{tabular}
\end{table}

\begin{table}[!h]
\centering
\small
\renewcommand{\arraystretch}{0.95}
\setlength{\tabcolsep}{2pt}

\begin{tabular}{p{0.05\linewidth} p{0.89\linewidth}}
\toprule
\multicolumn{2}{l}{\textbf{Algorithm 6}. Continual Alignment for SAM with MoDA} \\
\midrule
\multicolumn{2}{p{0.94\linewidth}}{\textbf{Input:} Pre-trained SAM $\theta$, Tasks $\mathcal{D}=\{D^{tr}_1,\ldots,D^{tr}_N\}$} \\
\multicolumn{2}{p{0.94\linewidth}}{\textbf{Output:} Alignment Layer Pool $\mathcal{P}=\{K_t:\,\Phi_t\}_{t=1}^N$, Task Classifier $\mathcal{T}$ (with global tokens $T\!\in\!\mathbb{R}^{L\times C}$)} \\
1:  & Initialize $\mathcal{P}\leftarrow\emptyset$, global tokens $T$, router $\mathcal{T}$. \\
2:  & \textbf{for} $t=1,\ldots,N$ \textbf{do} \\
3:  & \quad Initialize / load current alignment layer $\Phi_t$.\\
4:  & \quad \textbf{for} $(x,y)\in D^{tr}_t$ \textbf{do} \\
5:  & \quad\quad $Z=E_\theta(x)$;\ \ $\tilde Z=\Phi_t(Z)$;\ \ $\hat m=f_\theta(\tilde Z)$ \\
6:  & \quad\quad Update $\Phi_t$ by $\mathcal{L}_{align}(\hat{m}, y)$. \\
7:  & \quad \textbf{end for} \\
8:  & \quad Save $\Phi_t$ into pool: $\mathcal{P}\leftarrow \mathcal{P}\cup\{K_t:\Phi_t\}$. \\
9:  & \quad Select exemplars for Memory Bank $\mathcal{M}$. \\
10: & \quad \textbf{for} $x\in\mathcal{M}$ \textbf{do} \\
11: & \quad\quad global feature $q=f_{\theta}'(T)[0]$. \\
12: & \quad\quad Update router $\mathcal{T}$ with CE loss $\mathcal{L}_{ce}$. \\
13: & \quad \textbf{end for} \\
14: & \textbf{end for} \\
15: & \textbf{return}\ $\mathcal{P},\ \mathcal{T}$ \\
\bottomrule
\end{tabular}
\end{table}
\begin{table}[!h]
\centering
\small
\renewcommand{\arraystretch}{1.0}
\setlength{\tabcolsep}{2pt}

\begin{tabular}{p{0.04\linewidth} p{0.90\linewidth}}
\toprule
\multicolumn{2}{l}{\textbf{Algorithm 7}. Continual Alignment for SAM} \\
\midrule
\multicolumn{2}{l}{\textbf{Input:} Pre-trained SAM $\theta$, Tasks $\mathcal{D}=\{D^{tr}_1,\ldots,D^{tr}_N\}$} \\
\multicolumn{2}{l}{\textbf{Output:} Alignment Layer Pool $\mathcal{P}_{\mathcal{A}}=\{\mathcal{A}_t\}_{t=1}^N$,} \\
\multicolumn{2}{l}{VAE Router $\mathcal{P}_{\mathcal{V}}=\{\mathcal{V}_t\}_{t=1}^N$, Thresholds $\mathcal{T}=\{\tau_t\}_{t=1}^N$} \\

1: & Initialize $\mathcal{P}_{\mathcal{A}}\leftarrow\emptyset$, $\mathcal{P}_{\mathcal{V}}\leftarrow\emptyset$, $\mathcal{T}\leftarrow\emptyset$. \vspace{-3pt} \\
2: & \textbf{for} $t=1,\ldots,N$ \textbf{do}\\
3: & \quad \textbf{for} $(x,y) \in D^{tr}_t$ \textbf{do} \\
4: & \quad\quad $Z = E_\theta(x)$; $\tilde{Z} = \mathcal{A}_t(Z)$. $\hat{m} = D_\theta(\tilde{Z})$. \\
5: & \quad\quad Update $\mathcal{A}_t$ with loss $\mathcal{L}_{align}(\hat{m}, y)$. \\
6: & \quad \textbf{end for} \\
7: & \quad Train VAE $\mathcal{V}_t$ on $D^{tr}_t$ using $Z$. \\
8: & \quad Compute $\tau_t$ for $\mathcal{V}_t$ \\
9: & \quad Save $\mathcal{A}_t$, $\mathcal{V}_t$, $\tau_t$ into pools $\mathcal{P}_{\mathcal{A}}$, $\mathcal{P}_{\mathcal{V}}$, $\mathcal{T}$. \\
10: & \textbf{end for} \\
11: & \textbf{return} $\mathcal{P}_{\mathcal{A}}, \mathcal{P}_{\mathcal{V}}, \mathcal{T}$ \vspace{3pt} \\
\midrule
12: & \textbf{Inference Phase:} Given a test image $I_{test}$. \\
13: & \quad $Z_{test} = E_\theta(I_{test})$. \\
14: & \quad \textbf{for} $t=1,\ldots,N$ \textbf{do} \\
15: & \quad\quad Compute $s_t = \mathcal{L}_{VAE_t}(Z_{test})$. \\
16: & \quad \textbf{end for} \\
17: & \quad $t^* = \arg\min_t s_t$. \\
18: & \quad \textbf{if} $s_{t^*} < \tau_{t^*}$ \\
    & \quad\quad Load $\mathcal{A}_{t^*}$; $\tilde{Z}_{test} = \mathcal{A}_{t^*}(Z_{test})$.\\
19: & \quad \textbf{else} \\
20: & \quad\quad Load Identity Alignment Layer$\mathcal{A}^*$ \\ 
    & \quad\quad $\tilde{Z}_{test} = \mathcal{A}^*(Z_{test})$.\\
21: & \quad \textbf{end if} \\
22: & \quad $\hat{m}_{test} = D_\theta(\tilde{Z}_{test})$. \\
\bottomrule
\end{tabular}
\end{table}

\subsubsection{Evaluation on Task-Order Robustness}
According to~\Cref{tab:three-orders-continual-learning-results-med}, different continual learning methods exhibit clear variations in performance when the task order changes. Traditional continual learning approaches such as LwF, EWC, and naive adapter fine-tuning show large fluctuations in Last-IoU, Avg-IoU, and forgetting across different orders, indicating strong sensitivity to cross-task distribution shifts and a lack of robustness to task ordering. ER is relatively more stable, but its performance is still \mbox{affected by task permutations.}

In contrast, both MoDA and CA-SAM maintain consistently high performance under all three task orders. Since these routing-based methods activate separate adapter parameters for each task, they naturally avoid interference between tasks and are therefore largely invariant to task order. Notably, CA-SAM achieves the highest Last-IoU and Avg-IoU as well as the lowest forgetting across all orders, demonstrating the strongest task-order robustness.

Such insensitivity to task order is particularly important for real-world medical continual learning, where data from different hospitals, modalities, and anatomical regions rarely arrive in a fixed or predetermined sequence.

\definecolor{mycrimson}{HTML}{DC143C}   
\definecolor{mylavender}{HTML}{E6E6FA}  

\begin{table*}[t]
  \centering
  \begin{minipage}{\textwidth}
  \centering
  \setlength{\tabcolsep}{4pt}
  \renewcommand{\arraystretch}{1.15}

  \begin{tabular}{p{0.20\textwidth}|p{0.12\textwidth}p{0.12\textwidth}p{0.12\textwidth}|p{0.12\textwidth}p{0.12\textwidth}p{0.12\textwidth}}
    \toprule\toprule
    \multirow{2}{*}{\textbf{Method}} &
    \multicolumn{3}{c|}{\textbf{IoU on Med}} & \multicolumn{3}{c}
    {\textbf{BIoU on Med}}\\
             & Last-IoU \(\uparrow\) & Avg-IoU \(\uparrow\) & FF-IoU \(\downarrow\)
             & Last-BIoU \(\uparrow\) & Avg-BIoU \(\uparrow\) & FF-BIoU \(\downarrow\) \\
    \midrule
    \multicolumn{7}{c}{\textbf{DN} \(\rightarrow\) \textbf{56Nx} \(\rightarrow\) \textbf{EBHI-SEG} \(\rightarrow\) \textbf{STS-2D} \(\rightarrow\) \textbf{promise12} \(\rightarrow\) \textbf{MSD\_Spleen} \(\rightarrow\) \textbf{MSD\_Prostate} \(\rightarrow\) \textbf{Polyp} \(\rightarrow\) \textbf{ACDC}} \\
    \midrule
    SAM~\cite{kirillov2023segment} + AL(naive)    & 25.27 & 34.06 & 54.13\% & 25.15 & 25.94 & 37.12\% \\
    LwF~\cite{li2017learning}                     & 30.19 & 45.27 & 6.55\%   & 18.81 & 30.08 & 8.96\%  \\
    EWC~\cite{kirkpatrick2017overcoming}          & 25.38 & 31.01 & 46.68\%  & 19.30 & 23.00 & 32.44\% \\
    ER~\cite{rolnick2019experience}               & \cellcolor{mylavender!50}{73.73} & \cellcolor{mylavender!50}{74.68} & 6.20\%  & \cellcolor{mylavender!50}{51.03} & \cellcolor{mylavender!50}{53.00} & 11.50\% \\
    L2P~\cite{wang2022learning}                   & 67.65 & 70.48 & 2.47\%  & 41.71 & 43.78 & 4.68\%  \\
    MoDA~\cite{yang2024continual}                 
      & 65.83 & 67.02 & \cellcolor{mylavender!50}{2.20\% }
      & 49.75 & 48.67 & \cellcolor{mycrimson!18}{1.12\%}  \\
    CA-SAM (Ours)                                  
      &\cellcolor{mycrimson!18}\(\mathbf{75.61}\)
      & \cellcolor{mycrimson!18}\(\mathbf{75.38}\)
      & \cellcolor{mycrimson!18}\(\textbf{1.71\% }\)
      & \cellcolor{mycrimson!18}\(\mathbf{58.94}\)
      & \cellcolor{mycrimson!18}\(\mathbf{57.30}\)
      & \cellcolor{mylavender!50}\(\mathbf{1.84\%}\) \\

    \midrule
    \multicolumn{7}{c}{\textbf{EBHI-SEG} \(\rightarrow\) \textbf{Polyp} \(\rightarrow\) \textbf{ACDC} \(\rightarrow\) \textbf{MSD\_Prostate} \(\rightarrow\) \textbf{56Nx} \(\rightarrow\) \textbf{MSD\_Spleen} \(\rightarrow\) \textbf{DN} \(\rightarrow\) \textbf{promise12} \(\rightarrow\) \textbf{STS-2D}} \\
    \midrule
    SAM~\cite{kirillov2023segment} + AL(naive)    & 13.13 & 31.98 & 66.10\% & 12.01 & 24.14 & 49.55\% \\
    LwF~\cite{li2017learning}                     & 25.22 & 44.26 & 1.66\% & 12.99 & 26.49 & 2.98\%  \\
    EWC~\cite{kirkpatrick2017overcoming}          & 18.38 & 32.81 & 52.07\% & 12.29 & 24.50 & 38.97\% \\
    ER~\cite{rolnick2019experience}               & \cellcolor{mylavender!50}{68.57} & 68.19 & 10.92\% & 47.47 & 44.99 & 15.00\% \\
    MoDA~\cite{yang2024continual}                 
      & 67.03 & \cellcolor{mylavender!50}{71.97} & \cellcolor{mylavender!50}{1.44\%}
      & \cellcolor{mylavender!50}{51.44} & \cellcolor{mylavender!50}{51.71} & \cellcolor{mycrimson!18}{1.06\%}  \\
    CA-SAM (Ours)                                  
      &\cellcolor{mycrimson!18}\(\mathbf{75.78}\)
      & \cellcolor{mycrimson!18}\(\mathbf{76.47}\)
      & \cellcolor{mycrimson!18}\(\mathbf{1.31\%}\)
      & \cellcolor{mycrimson!18}\(\mathbf{57.73}\)
      & \cellcolor{mycrimson!18}\(\mathbf{54.79}\)
      & \cellcolor{mylavender!50}\(\mathbf{1.64\%}\) \\

    \midrule
\multicolumn{7}{c}{\textbf{MSD\_Prostate} \(\rightarrow\) \textbf{56Nx} \(\rightarrow\) \textbf{STS-2D} \(\rightarrow\) \textbf{Polyp} \(\rightarrow\) \textbf{DN} \(\rightarrow\) \textbf{MSD\_Spleen} \(\rightarrow\) \textbf{ACDC} \(\rightarrow\) \textbf{EBHI-SEG} \(\rightarrow\) \textbf{promise12}} \\
\midrule
SAM~\cite{kirillov2023segment} + AL(naive)  & 28.39 & 36.85 & 52.76\% & 19.54 & 29.60 & 45.40\% \\
LwF~\cite{li2017learning}  & 13.43 & 20.58 & 1.76\% & 9.84 & 16.43 & 1.86\% \\
EWC~\cite{kirkpatrick2017overcoming}  & 22.99 & 37.88 & 45.78\% & 17.83 & 28.00 & 27.91\% \\
ER~\cite{rolnick2019experience}  & \cellcolor{mylavender!50}{71.69} & \cellcolor{mylavender!50}{62.13} & 7.76\% & \cellcolor{mylavender!50}{53.11} & \cellcolor{mylavender!50}{46.74} & 7.94\% \\
MoDA~\cite{yang2024continual}
& 67.26 & 54.03 & \cellcolor{mycrimson!18}{0.96\%}
& 51.54 & 42.38 & \cellcolor{mycrimson!18}{0.90\%} \\
CA-SAM (Ours)
    &\cellcolor{mycrimson!18}\(\mathbf{76.19}\)
    & \cellcolor{mycrimson!18}\(\mathbf{63.31}\)
    & \cellcolor{mylavender!50}\(\mathbf{1.66\%}\)
    & \cellcolor{mycrimson!18}\(\mathbf{59.51}\)
    & \cellcolor{mycrimson!18}\(\mathbf{52.74}\)
    & \cellcolor{mylavender!50}\(\mathbf{1.70\%}\) \\
    \bottomrule\bottomrule
  \end{tabular}

  \caption{The results of different contrastive methods under three different task orders on medical datasets.The
    \colorbox{mycrimson!18}{best}
     and 
    \colorbox{mylavender!50}{second best}
    performances are highlighted.
  }
  \label{tab:three-orders-continual-learning-results-med}
  \end{minipage}
\end{table*}
\subsubsection{The Confidence Threshold $\tau_t$ for Each Task}
\Cref{tab:dataset_thresholds} presents the calibrated thresholds $\tau_t$ discussed in the router threshold ablation study. These thresholds serve as the critical decision boundaries for the VAE Router to distinguish between known tasks and out-of-distribution (OOD) samples. As detailed in the implementation settings, these values were derived via 5-fold cross-validation on the training set of each task based on the in-distribution ELBO score distribution. We report the thresholds calculated using four different statistical criteria: $\mu+2\sigma$, $p_{95}$, $p_{97}$, and $p_{99}$. For CA-SAM, we adopt the $p_{97}$ strategy, which was experimentally determined to offer the optimal trade-off between preserving known-task segmentation performance and effectively rejecting unseen domains.
\begin{table*}[!h]
  \centering
  \small 
  \begin{tabular}{cccccccccc}
    \toprule
    Parameter & ACDC & EBHI-SEG & 56Nx & DN & Polyp & MSD-Prostate & MSD-Spleen & Promise12 & STS-2D \\
    \midrule
    $\mu + 2\sigma$ & 0.0790 & 0.0680 & 0.2121 & 0.1589 & 0.1462 & 0.1857 & 0.1257 & 0.1638 & 0.0632 \\
    $p_{95}$ & 0.0761 & 0.0653 & 0.2066 & 0.1553 & 0.1446 & 0.1817 & 0.1199 & 0.1404 & 0.0622 \\
    $p_{97}$(Ours) & \textbf{0.0813} & \textbf{0.0690} & \textbf{0.2258} & \textbf{0.1646} & \textbf{0.1494} & \textbf{0.1915} & \textbf{0.1285} & \textbf{0.1483} & \textbf{0.0662} \\
    $p_{99}$ & 0.0916 & 0.0806 & 0.2808 & 0.1819 & 0.1766 & 0.2062 & 0.1544 & 0.1782 & 0.0729 \\
    \bottomrule
  \end{tabular}
  \centering
  \caption{Calibrated Thresholds ($\tau_t$) for Each Dataset under Different Strategies.}
  \label{tab:dataset_thresholds}
\end{table*}

\subsection{Explainability Metrics Computation}
\label{sec:Appendix-Explainability}

We selected the TV (Total Variation) and JS (Jensen-Shannon) divergence metrics for comparison.
The two metrics are computed as follows:

\begin{equation}
D_{TV}(P, Q) = \frac{1}{2} \sum_{x} |P(x) - Q(x)| 
\end{equation}
\begin{equation}
D_{JS}(P, Q) = \frac{1}{2} \left( D_{KL}(P \parallel M) + D_{KL}(Q \parallel M) \right)
\end{equation}

where $M = \frac{1}{2}(P + Q)$ is the average of the two distributions, and $D_{KL}$ is the KL divergence.

\subsection{Ablation Study}


\subsubsection{Cross-dataset Stability Ablation}

\noindent\textbf{Temperature Ablation of Attention Pooling.}
The temperature parameter $T$ in the attention pooling mechanism controls the smoothness of the Softmax function used for spatial feature aggregation. To evaluate the sensitivity of our framework to this hyperparameter, we conducted an ablation study by varying $T$ from 0.1 to 16, as detailed in \Cref{tab:ablation_temp}. The experimental results demonstrate that our VAE Router exhibits strong robustness to variations in temperature. Across a wide range of values ($T \in [1, 16]$), the model consistently maintains high task identification performance, with Zero-shot Accuracy stabilizing above 97\% and Average IoU remaining steady. Therefore, since the framework exhibits such strong robustness within this broad range, we employ the standard and computationally simple setting of $T=1$ as the default configuration in our main experiments, prioritizing model simplicity \mbox{without sacrificing performance.}

\begin{table*}[t]
  \centering
  \begin{tabular}{cccccccccccc}
    \toprule
    \multirow{2}{*}{Parameter} & \multicolumn{3}{c}{IoU on Med} & \multicolumn{3}{c}{BIoU on Med} & \multicolumn{5}{c}{Zero-shot} \\
    \cmidrule(lr){2-4} \cmidrule(lr){5-7} \cmidrule(lr){8-12}
     & Last & Avg & FF & Last & Avg & FF & IoU & BIoU & Acc & IoU(Med) & BIoU(Med) \\
    \midrule
    $T=0.1$ & 76.04 & 76.79 & 1.58\% & 60.05 & 59.55 & 0.27\% & 55.68 & 58.15 & 99.03 & 76.07 & 59.99 \\
    $T=0.25$ & 75.85 & 76.77 & 1.62\% & 59.92 & 59.52 & 0.13\% & 55.99 & 58.33 & 99.49 & 75.93 & 59.83 \\
    $T=0.5$ & 75.54 & 76.55 & 1.87\% & 60.04 & 59.54 & 0.14\% & 55.24 & 57.90 & 98.36 & 75.91 & 60.03 \\
    $T=1$ & 75.73 & 76.75 & 1.78\% & 59.80 & 59.51 & 0.29\% & 54.99 & 57.74 & 97.86 & 75.62 & 59.57 \\
    $T=2$ & 76.05 & 76.88 & 1.32\% & 59.48 & 59.21 & 0.38\% & 55.72 & 58.14 & 99.06 & 76.41 & 59.79 \\
    $T=4$ & 76.07 & 76.97 & 1.42\% & 59.74 & 59.41 & 0.35\% & 55.34 & 57.88 & 98.42 & \textbf{76.46} & 59.96 \\
    $T=8$ & \textbf{76.24} & \textbf{76.99} & \textbf{1.23\%} & 59.82 & 59.39 & 0.20\% & \textbf{56.00} & \textbf{58.34} & \textbf{99.53} & 76.38 & 59.76 \\
    $T=16$ & 75.78 & 76.75 & 1.68\% & \textbf{60.07} & \textbf{59.57} & \textbf{0.07\%} & 55.47 & 58.02 & 98.68 & 76.12 & \textbf{60.19} \\
    \bottomrule
  \end{tabular}
  \centering
  \caption{Ablation Study on Attention Pooling Temperature Parameter $T$}
  \label{tab:ablation_temp}
\end{table*}

\renewcommand{\arraystretch}{0.98}

\begin{table*}[ht]
  \centering
  
  \begin{tabular}{cccccccccccc}
    \toprule
    \multirow{2}{*}{Parameter} & \multicolumn{3}{c}{IoU on Med} & \multicolumn{3}{c}{BIoU on Med} & \multicolumn{5}{c}{Zero-shot} \\
    \cmidrule(lr){2-4} \cmidrule(lr){5-7} \cmidrule(lr){8-12}
     & Last & Avg & FF & Last & Avg & FF & IoU & BIoU & Acc & IoU(Med) & BIoU(Med) \\
    \midrule
    $\beta=0 $   & 44.81 & 52.55 & 4.25\% & 36.74 & 42.08 & 3.55\% & 46.64 & 53.56 & 84.59 & 53.25 & 44.14 \\
    $\beta=1$    & 44.64 & 49.73 & 5.17\% & 36.39 & 40.18 & 4.51\% & 36.69 & 41.11 & 14.07 & 44.24 & 36.60 \\
    $\beta=1.5$  & 48.35 & 55.01 & 6.12\% & 39.44 & 44.10 & 4.73\% & 36.88 & 41.23 & 11.62 & 52.32 & 42.37 \\
    $\beta=2$    & 58.75 & 62.70 & 3.43\% & 47.06 & 49.49 & 2.96\% & 39.66 & 44.41 & 29.36 & 62.12 & 49.91 \\
    $\beta=2.5$  & 64.26 & 67.03 & 3.15\% & 51.49 & 52.69 & 2.37\% & 44.42 & 49.75 & 56.18 & 66.18 & 53.09 \\
    $\beta=3$    & 68.48 & 70.46 & 1.96\% & 53.75 & 54.50 & 1.64\% & 49.59 & 53.99 & 84.14 & 71.75 & 56.24 \\
    $\beta=3.5$  & 68.80 & 70.17 & 1.22\% & 54.19 & 54.69 & 1.26\% & 48.49 & 52.82 & 72.24 & 71.70 & 56.81 \\
    $\beta=4$    & 69.52 & 71.37 & 1.35\% & 55.50 & 55.89 & 0.97\% & 51.14 & 55.52 & 88.42 & 72.40 & 57.56 \\
    $\beta=4.5$  & 73.13 & 74.75 & 0.82\% & 57.51 & 57.80 & 0.61\% & 52.86 & 56.74 & 94.15 & 74.54 & 58.54 \\
    $\beta=5$    & 72.79 & 74.44 & 1.33\% & 57.53 & 57.71 & 0.77\% & 52.68 & 56.75 & 94.30 & 75.18 & 59.49 \\
    $\beta=5.5$  & 73.19 & 74.41 & \textbf{0.51\%} & 58.01 & 57.89 & 0.40\% & 53.56 & 57.20 & 95.58 & 74.44 & 58.78 \\
    $\beta=6$    & 73.54 & 74.99 & 0.80\% & 57.97 & 58.21 & 0.85\% & 54.13 & 57.51 & 96.52 & 74.98 & 59.40 \\
    $\beta=6.5$  & 74.43 & 75.86 & 1.72\% & 58.76 & 58.85 & 0.57\% & 53.74 & 57.23 & 95.88 & 76.08 & 60.00 \\
    $\beta=7$    & 73.86 & 75.25 & 0.87\% & 58.46 & 58.42 & 0.68\% & 55.48 & 58.09 & 98.63 & 75.61 & 59.79 \\
    $\beta=7.5$  & 74.26 & 75.46 & 0.69\% & 58.52 & 58.58 & 0.55\% & 55.11 & 57.85 & 98.00 & 75.40 & 59.43 \\
    $\beta=8$    & 75.25 & 76.25 & 1.14\% & 59.14 & 58.86 & 0.39\% & 52.79 & 55.62 & 93.04 & 75.70 & 59.46 \\
    $\beta=8.5$  & 75.72 & 76.47 & 0.75\% & 59.37 & 58.99 & 0.30\% & 55.77 & 58.23 & 99.09 & \textbf{76.34} & 59.95 \\
    $\beta=9 $   & 75.06 & 76.24 & 1.64\% & 59.11 & 59.02 & 0.51\% & 55.34 & 57.96 & 98.48 & 75.84 & 59.67 \\
    $\beta=9.5$  & 75.04 & 76.17 & 1.06\% & 59.45 & 59.14 & 0.32\% & 55.73 & 58.18 & 99.09 & 75.75 & 59.77 \\
    $\beta=10$   & 75.76 & 76.78 & 1.60\% & 59.58 & 59.24 & 0.27\% & 55.55 & 58.11 & 98.69 & 76.25 & 59.93 \\
    $\beta=10.5$ & 75.41 & 76.38 & 1.53\% & 59.60 & 59.34 & 0.40\% & 55.73 & 58.18 & 99.02 & 76.02 & 60.01 \\
    $\beta=11$   & 75.66 & 76.64 & 1.53\% & 59.72 & 59.30 & 0.23\% & 55.41 & 58.11 & 98.70 & 75.98 & 59.80 \\
    $\beta=11.5$ & 75.25 & 76.33 & 1.73\% & 59.61 & 59.32 & 0.39\% & 55.88 & 58.27 & 99.32 & 76.20 & \textbf{60.07} \\
    $\beta=12$   & 75.60 & 76.61 & 1.49\% & 59.46 & 59.22 & 0.38\% & 56.01 & 58.36 & 99.53 & 75.77 & 59.60 \\
    $\beta=12.5$ & 75.48 & 76.81 & 2.03\% & 59.41 & 59.33 & 0.55\% & 54.85 & 57.78 & 97.84 & 75.75 & 59.43 \\
    $\beta=13$   & 75.42 & 76.46 & 1.40\% & 59.60 & 59.30 & 0.27\% & 54.99 & 57.88 & 97.99 & 76.06 & 60.03 \\
    $\beta=13.5$ & 75.73 & 76.74 & 1.73\% & 59.78 & 59.49 & 0.32\% & 55.65 & 58.11 & 98.98 & 76.04 & 60.03 \\
    $\beta=14$   & 75.81 & 76.85 & 1.56\% & 59.46 & 59.25 & 0.36\% & 55.95 & 58.26 & 99.41 & 75.92 & 59.53 \\
    $\beta=14.5$ & 75.81 & 76.88 & 1.66\% & 59.70 & 59.45 & 0.27\% & 56.01 & 58.34 & 99.57 & 75.99 & 59.57 \\
    $\beta=15$   & 75.44 & 76.58 & 1.84\% & 59.70 & 59.49 & 0.35\% & \textbf{56.18} & \textbf{58.44} & \textbf{99.79} & 75.91 & 59.98 \\
    $\beta=15.5$ & 75.77 & 76.69 & 1.82\% & 59.94 & 59.51 & 0.22\% & 55.71 & 58.19 & 99.11 & 75.68 & 59.64 \\
    $\beta=16$   & 76.05 & 76.90 & 1.47\% & 59.98 & \textbf{59.56} & 0.17\% & 55.42 & 57.91 & 98.66 & 75.87 & 59.86 \\
    $\beta=16.5$ & \textbf{76.12} & \textbf{76.90} & 1.43\% & 59.95 & 59.45 & 0.24\% & 55.63 & 58.13 & 99.05 & 76.23 & 59.89 \\
    $\beta=17$   & 76.00 & 76.90 & 1.37\% & 59.83 & 59.50 & 0.27\% & 55.98 & 58.27 & 99.41 & 76.14 & 59.89 \\
    $\beta=17.5$ & 75.88 & 76.68 & 1.71\% & \textbf{60.03} & 59.52 & \textbf{0.05\%} & 55.85 & 58.23 & 99.30 & 75.78 & 59.81 \\
    $\beta=18$   & 75.58 & 76.56 & 1.73\% & 59.74 & 59.38 & 0.21\% & 55.97 & 58.33 & 99.48 & 75.71 & 59.71 \\
    \bottomrule
  \end{tabular}
   \centering
  \caption{Ablation study on the KL regularization coefficient $\beta$ in the VAE Router.}
  \vspace{-30pt}
  \label{tab:VAE_structure_ablation}
\end{table*}

\noindent\textbf{VAE Structure Ablation.}
To investigate the role of the KL divergence in our task routing mechanism, we conducted a sensitivity analysis on the regularization coefficient $\beta$ in our ELBO loss, varying it from 0 to 18. As shown in \Cref{tab:VAE_structure_ablation}, $\beta$ is pivotal in balancing feature reconstruction and the constraint on the latent space distribution. In the lower range ($\beta < 6$), the model exhibits significant instability. As $\beta$ increases, the stronger KL penalty drives the formation of more separated and structured distribution boundaries for each task in the latent space. This consequently enforces a distinct and compact feature distribution for every task, which significantly enhances the discriminability between different tasks and is crucial for the router to reject OOD samples. We observe a substantial performance stabilization for $\beta \in [7, 18]$, where the Average IoU on Med consistently stays above 75\% and \mbox{Zero-shot Accuracy exceeds 98\%.}

\end{document}